\documentclass[journal]{IEEEtran}
\usepackage{lmodern}
\usepackage{amssymb}
\usepackage{xcolor}
\usepackage[left=2.00cm, right=2.00cm, top=2.50cm, bottom=2.00cm]{geometry}
\usepackage{hyperref}
\hypersetup{
	colorlinks=true,
	linkcolor=.,
	filecolor=magenta,      
	urlcolor=cyan,
	bookmarks=true,
}
\usepackage{subcaption}
\usepackage{algorithm}
\usepackage{algorithmic}

\usepackage{enumitem}
\setlist[itemize]{left=0pt}
\setlist[enumerate]{left=0pt}
\newcommand{\uvec}[1]{\underline{#1}\vphantom{#1}}
\newcommand{\duvec}[1]{\uvec{\dot{#1}}}
\newcommand{\dduvec}[1]{\uvec{\ddot{#1}}}
\newcommand{\uoh}{\uvec{0}}

\newcommand{\ua}{\uvec{a}}
\newcommand{\ub}{\uvec{b}}
\newcommand{\uf}{\uvec{f}}
\newcommand{\ug}{\uvec{g}}
\newcommand{\uh}{\uvec{h}}
\newcommand{\ui}{\uvec{i}}
\newcommand{\uj}{\uvec{j}}
\newcommand{\uk}{\uvec{k}}
\newcommand{\up}{\uvec{p}}
\newcommand{\uq}{\uvec{q}}
\newcommand{\duq}{\duvec{q}}
\newcommand{\dduq}{\dduvec{q}}
\newcommand{\uu}{\uvec{u}}
\newcommand{\uv}{\uvec{v}}
\newcommand{\ux}{\uvec{x}}

\newcommand{\uF}{\uvec{F}}

\newcommand{\uM}{\uvec{M}}
\newcommand{\uV}{\uvec{V}}
\newcommand{\duV}{\duvec{V}}
\newcommand{\uW}{\uvec{W}}

\newcommand{\uphi}{\uvec{\phi}}
\newcommand{\duphi}{\duvec{\phi}}
\newcommand{\uom}{\uvec{\omega}}

\newcommand{\pt}[1]{\text{\textsc{#1}}}

\newcommand{\pE}{\pt{e}}
\newcommand{\pF}{\pt{f}}
\newcommand{\pG}{\pt{g}}

\newcommand{\pK}{\pt{k}}
\newcommand{\pO}{\pt{o}}
\newcommand{\pP}{\pt{p}}

\newcommand{\pW}{\pt{w}}
\newcommand{\pZ}{\pt{z}}

\newcommand{\vv}{\vec{v}}
\newcommand{\vi}{\vec{\imath}}
\newcommand{\vj}{\vec{\jmath}}
\newcommand{\vom}{\vec{\omega}}

\newcommand{\bR}{\mathbb{R}} 
\newcommand{\cI}{\mathcal{I}}
\newcommand{\cN}{\mathcal{N}}
\newcommand{\cW}{\mathcal{W}}

\usepackage[mathscr]{euscript}
\newcommand{\sP}{\mathscr{P}}

\newcommand{\SdR}[1]{\fbox{$ #1 $}}

\newcommand{\atandue}[1]{\mathop{\mathrm{atan2}}\di{#1}}

\newcommand{\roll}{\varphi}
\newcommand{\pitch}{\theta}
\newcommand{\yaw}{\psi}

\newcommand{\ads}{\mathop{\mathrm{ad}^*}}
\newcommand{\I}{\cI} 

\newcommand{\matrice}[1]{\begin{bmatrix}
		#1
\end{bmatrix}}
\newcommand{\rank}{\mathop{\mathrm{rank}}}
\newcommand{\diag}[1]{\mathrm{diag}\graffe{#1}}

\newcommand{\tonde}[1]{\left( #1 \right)}
\newcommand{\quadre}[1]{\left[ #1 \right]}
\newcommand{\graffe}[1]{\left\lbrace #1 \right\rbrace}
\newcommand{\di}[1]{\! \tonde{#1}}

\newcommand{\op}[1]{\!\quadre{#1}}

\newcommand{\bigtonde}[1]{\,\big( #1 \big)}
\newcommand{\bigquadre}[1]{\,\big[#1\big]}

\newcommand{\newlineparentheses}{\right.\\ \left.}

\newcommand{\abs}[1]{ \left\lvert{#1}\right\rvert}
\newcommand{\norm}[1]{ \left\lVert #1 \right\rVert}
\newcommand{\bignorm}[1]{ \big\lVert #1 \big\rVert}
\newcommand{\calcbw}[3]{\left.{#1}\right\rvert_{#2}^{#3}}

\newcommand{\oo}{\infty}
\newcommand{\tc}{\ \vert \ }
\newcommand{\forevery}{\ \forall \, }
\newcommand{\diff}[2]{\frac{\partial {#1}}{\partial {#2}}}
\usepackage{mathtools}
\newcommand{\est}{\coloneqq}

\newcommand{\THEN}{\Rightarrow}
\newcommand{\wt}[1]{\widetilde{#1}\vphantom{#1}}
\newcommand{\dwt}[1]{\dot{\widetilde{#1}}\vphantom{#1}}

\date{}
\author{
	Aristide Emanuele Casucci\textsuperscript{1},
	Federico Nesti\textsuperscript{1},
	Mauro Marinoni\textsuperscript{1},
	Giorgio Buttazzo\textsuperscript{1}\thanks{\textsuperscript{1}Department of Excellence on Robotics and AI, Sant'Anna School of Advanced Studies, Pisa, Italy}
}

\newcommand{\NAME}{\textsc{Feelbert}}
\title{Feelbert: A Feedback Linearization-based Embedded Real-Time Quadrupedal Locomotion Framework}

\begin{document}
	\maketitle
	\begin{abstract}

Quadruped robots have become quite popular for their ability to adapt their locomotion to generic uneven terrains.
For this reason, over time, several frameworks for quadrupedal locomotion have been proposed, but with little attention to ensuring a predictable timing behavior of the controller.

To address this issue, this work presents \NAME, a modular control framework for quadrupedal locomotion suitable for execution on an embedded system under hard real-time execution constraints. 
It leverages the feedback linearization control technique to obtain a closed-form control law for the body, valid for all configurations of the robot. 
The control law was derived after defining an appropriate rigid body model that uses the accelerations of the feet as control variables, instead of the estimated contact forces. 
This work also provides a novel algorithm to compute footholds and gait temporal parameters using the concept of imaginary wheels, and a heuristic algorithm to select the best gait schedule for the current velocity commands.

The proposed framework is developed entirely in C++, with no dependencies on third-party libraries and no dynamic memory allocation, to ensure predictability and real-time performance.
Its implementation allows \NAME\ to be both compiled and executed on an embedded system for critical applications, as well as integrated into larger systems such as Robot Operating System 2 (ROS 2).
For this reason, \NAME\ has been tested in both scenarios, demonstrating satisfactory results both in terms of reference tracking and temporal predictability, whether integrated into ROS 2 or compiled as a standalone application on a Raspberry Pi 5.

\end{abstract}
	\section{Introduction}

In recent years, researchers have put increasing effort in creating novel solutions for legged robots due to their ability to adapt their locomotion to generic uneven terrains.
Robots with at least six legs can easily achieve a stable walk by always keeping at least three feet on the ground, but they are more expensive to build.
On the other hand, the locomotion of quadrupedal and bipedal robots requires going through moments of instability, while they have less than three feet on the ground.
Therefore, they are the most studied among legged robots.
	
Over the years, various solutions have been proposed to address this problem, generally considering simplified kinematic models, such as spring-loaded inverted pendulum (SLIP)~\cite{tang2017stable,akbas2012zero}, or linearized dynamic models controlled with optimal control techniques such as model predictive control (MPC)~\cite{farshidian2017efficient,norby2022quad, ding2021representation,ding2019real,hamed2020quadrupedal,risbourg2022real}, or whole-body control (WBC)~\cite{risbourg2022real, raiola2020simple,bellicoso2017dynamic,li2023real}.
More recent approaches involve neural controllers such as central pattern generator (CPG)~\cite{barasuol2013reactive,zhang2023synloco} and reinforcement learning (RL)~\cite{zhang2023synloco, tan2021hierarchical}.

Each of these publicly available solutions presents its advantages, but none of them has been designed with the intent of being deployed on an embedded system with hard real-time execution constraints.
To the best of our knowledge, only a few works considered real-time execution of their framework, but only in terms of achieving relatively high execution frequencies on their platform, without providing timing guarantees~\cite{norby2022quad,ding2019real,hamed2020quadrupedal,risbourg2022real,raiola2020simple,bellicoso2017dynamic,li2023real}.
However, respecting hard real-time execution constraints is crucial for safety-critical applications where predictability is paramount and computational resources are limited.

\subsection{Method Overview} \label{ssec:overview}

Since quadrupedal locomotion is a complex problem, it is typically addressed by dividing it into sub-problems.
For this reason, many works proposed modular frameworks~\cite{norby2022quad,hamed2020quadrupedal,risbourg2022real,raiola2020simple,li2023real,barasuol2013reactive,tan2021hierarchical,fawcett2021real,fahmi2019passive,mastalli2020motion,rathod2021model}, in which each module solves a sub-problem.
Some of such frameworks are divided into a gait planning module, a body planning module, and a body tracking module~\cite{hamed2020quadrupedal,li2023real,barasuol2013reactive,fawcett2021real,mastalli2020motion,rathod2021model}.

\NAME\ follows a similar modular architecture, but it is the first quadruped locomotion framework designed to ensure a predictable execution behavior suitable for implementation on embedded systems with limited resources and hard real-time execution constraints.
This is achieved thanks to the adoption of simple algorithms to obtain lower execution times, improve temporal predictability, and reduce the memory footprint.
The proposed implementation also avoids the use of third-party libraries to improve timing predictability, adaptability, and maintainability.

\begin{figure*}[t]
	\centering
	\includegraphics[width=0.8\linewidth]{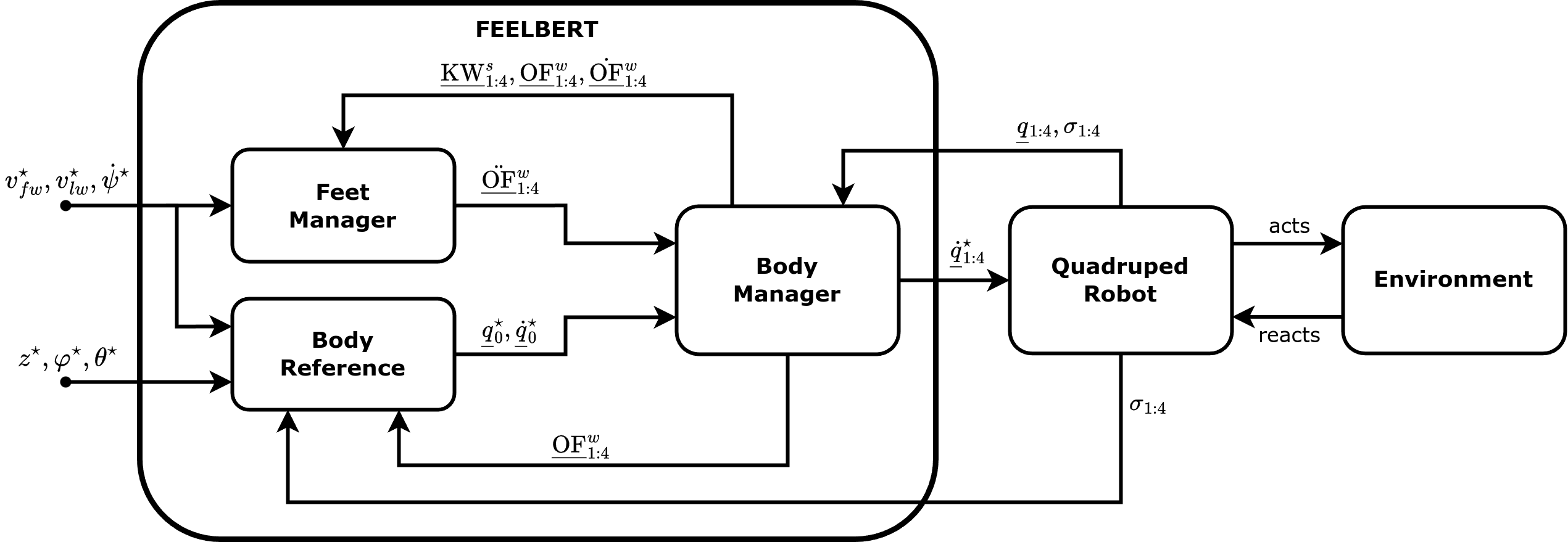}
	\caption{\NAME\ block diagram.}
	\label{fig:fwbd}
\end{figure*}

Figure~\ref{fig:fwbd} summarizes \NAME's architecture.
The framework receives high-level velocity commands ($v_{fw}^\star$, $v_{lw}^\star$, $\dot{\psi}^\star$) and height and inclination commands ($z^\star$, $\varphi^\star$, $\theta^\star$) from a joystick or a motion planner.
The main feedback from the robot is defined by the foot contact signals $\sigma_{1:4}$, and the joint angle values $\uq_{1:4}$.
The framework outputs the joint velocity commands $\duq_{1:4}^\star$ to move the legs.
The framework is organized in 3 macro-blocks: \textit{Feet Manager}, \textit{Body Reference}, and \textit{Body Manager}.

The \textit{Feet Manager} is responsible for converting the high-level velocity commands into the trajectory of each foot, by first computing the footholds, the gait schedule, and the gait period.
It outputs the feet control signals, which are sent to the \textit{Body Manager}.

The \textit{Body Reference} translates high-level commands and foot contact signals $\sigma_{1:4}$ into online references for the body and sends them to the \textit{Body Manager}.
Such references stabilize the body, by maintaining the Zero Moment Point (ZMP) within the Support Polygon.

The \textit{Body Manager} is responsible for low-level body control. Specifically, it receives the foot control signals from the \textit{Feet Manager}, the body references from the \textit{Body Reference}, and the feedback from the robot, returning the joint velocity commands $\duq_{1:4}^\star$ as output.
It also computes additional model data required by the \textit{Feet Manager} and the \textit{Body Reference} that will be specified in Section~\ref{ssec:bodyman}.

In the current implementation of \NAME, the \textit{Feet Manager} and \textit{Body Reference} blocks were designed assuming locomotion on a flat horizontal terrain.
In the future, these blocks will be extended to handle other types of terrain.
In any case, the modular division allows straightforward replacement or modification of these blocks, if required by the application.
\subsection{Contributions} \label{ssec:contributions}

The main contribution presented in this article is an efficient and predictable quadrupedal locomotion framework (\NAME), specifically designed to be executed on embedded boards with limited computational resources.
To achieve this goal, feedback linearization was applied to derive a single closed-form control law for the body, as detailed in Section~\ref{sssec:bodyctrl}.
To the best of our knowledge, this is the first work that applies a feedback linearization technique to the control of a quadruped robot, achieving satisfactory results both in terms of reference tracking and temporal predictability, as shown by the experiments in Section~\ref{sec:results}.
This control law holds for all the 15 cases in which the robot has at least one foot on the ground, thanks to the incorporation of the foot contact signals into the model of the robot to be controlled.
Such foot contact signals are independent of foot positions, making the model suitable for any type of terrain and walking.

The proposed model, detailed in Section~\ref{sssec:bodymodel}, adopts a new form of state space that uses foot accelerations as control inputs, while typical implementations use joint torques and estimated foot contact forces.
Furthermore, such a model is obtained by solving the inverse dynamics of the robot through LDQ matrix decomposition~\cite{golub2013matrix}, which always requires a fixed number of operations to execute, making it temporally predictable, as detailed in Section~\ref{sssec:ldq}.

In addition, the following improvements were introduced in gait management, always focusing on predictability and efficiency:
\begin{itemize}
	\item The projection of the main body onto the horizontal plane was modeled as a planar vehicle with pivoting wheels. 
	This allows 
	(i) computing online the footholds as a function of the velocity commands, as detailed in Section~\ref{sssec:wheelcmds}, and
	(ii) computing online the gait temporal parameters that maintain the final footholds within the foot workspaces, while keeping the feet off the ground for the shortest possible time, as detailed in Section~\ref{sssec:period}.
	\item A heuristic algorithm, described in Section~\ref{sssec:gaitsch}, is proposed to select the correct gait schedule, among the six different possibilities.
	It maximizes the longitudinal stability margin in each direction of motion (forward, backward, leftward, rightward, clockwise, counterclockwise) by determining the predominant motion in case of intermediate directions.
	\item Despite the common assumption that the foot remains stationary while touching the ground, other works neglected to choose a foot reference trajectory with both zero velocity and zero acceleration at the moment the foot touches the ground~\cite{raiola2020simple,li2023real,barasuol2013reactive,tan2021hierarchical,shi2021structural}.
	In this work, to fully respect the stationary foot assumption, foot reference trajectories are chosen to have both zero velocity and zero acceleration when the foot touches the ground.
\end{itemize}

\subsection{Outline}

The rest of the paper is organized as follows:
Section~\ref{sec:relwork} presents an overview of the existing approaches;
Section~\ref{sec:model} precisely formalizes the problem addressed in this work;
Section~\ref{sec:framework} illustrates the proposed solution;
and Section~\ref{sec:results} presents the results of the simulation experiments that have been carried out to show the effectiveness of the proposed approach.
	\section{Related Work} \label{sec:relwork}

Quadruped locomotion is a complex problem that is generally divided into two main sub-tasks: \textit{ body management} and \textit{foot management}.
The next two sections analyze the literature related to both tasks.

\subsection{Body Management}

This task consists of finding how to control the supporting legs in order to move the robot body in a desired direction while guaranteeing the equilibrium (i.e., without causing it to fall).
This is done by first modeling the robot body and then applying a specific control strategy to that model.
Therefore, body management is divided into two sub-problems: body modeling and body control.

\subsubsection{Body Modeling} \label{sssec:models}

Over the years, several models of increasing complexity have been proposed for legged robots, 
each having a different degree of accuracy in representation depending on the application scope.
The most well-known models proposed in the literature are presented below.
Typically, for a legged robot, the instantaneous center of rotation in the vertical plane is known as the \textit{Zero Moment Point} (ZMP)~\cite{kajita2008legged}.

Given that, the simplest model is the linear inverted pendulum (LIP)~\cite{tang2017stable}, which describes the body as an inverted pendulum composed of a point with a given mass at the top of a variable-length rod rotating around the ZMP.
The rod changes its length to keep the mass at a constant height while this moves.
This model is mainly used to compute the trajectory of the center of mass under the effect of gravity when a foot touches the ground, and therefore neglects the rotational dynamics of the robot's body.
This model was designed for biped robots~\cite{kajita2008legged} but was also applied to quadrupeds~\cite{tang2017stable, akbas2012zero, hamed2020quadrupedal, bellicoso2017dynamic, fawcett2021real}.

Since in many quadruped robots the masses of the leg links are negligible compared to that of the main body, the single rigid body dynamics (SRBD) model assumes that the robot's legs are massless, so that all the dynamic properties of the robot coincide with those of the central rigid body.
This is a fair approximation, thus it has been used in many works~\cite{farshidian2017efficient, norby2022quad, ding2021representation, ding2019real, li2023real, barasuol2013reactive, dicarlo2018dynamic, rathod2021model, chi2022linearization,farshidian2017real}.
This model has also been adopted in this work as presented in Section~\ref{sec:model}.

The full dynamic model (FDM) is the standard model of equations of motion for a robot, which takes into account the dynamic properties of all the robot's links. 
It is constructed using Lagrange dynamic equations or the Recursive Newton-Euler Algorithm (RNEA).
Since this model does not neglect any rigid body, it is highly accurate and it is employed in many works~\cite{farshidian2017efficient,raiola2020simple,bellicoso2017dynamic,li2023real,buchli2009compliant,fahmi2019passive,mastalli2020motion,farshidian2017robust}.
On the other hand, it is also quite complex to calculate manually.
Therefore, third-party libraries are typically used to compute the model components numerically.

\subsubsection{Body Control}

As reported by Chai et al.~\cite{chai2022survey}, the existing methods developed for motion control can be classified into two main categories: model-based methods and model-free methods.
%
Model-based control methods encompass various techniques that depend on the body model chosen from those listed in Section~\ref{sssec:models}.

The stability criterion control uses the LIP model to calculate a stable trajectory for the quadruped's center of mass or the ZMP.
The method ensures that a certain stability margin is met~\cite{igarashi2006free}, requiring detailed planning of foot and center of mass trajectories, thus sacrificing dynamic performance~\cite{chai2022survey}.

Inverse Dynamics Control employs a full dynamic model of a generic floating-base robot to compute the joint torques needed to follow the desired joint trajectories.
However, this approach does not address the problem of finding the desired joint trajectories required to move the quadruped robot, which must be solved separately.
This control technique has been applied to a quadruped robot by Buchli~\cite{buchli2009compliant} and Hutter~\cite{hutter2014quadrupedal}.
Buchli complemented it with an algorithm that plans the center of mass trajectories in advance.
Hutter complemented it by defining a hierarchical least squares optimization problem, where the optimization variables are the joint torques, joint accelerations, and foot contact forces.

Model predictive control (MPC) is an optimal control technique that uses a linear model of a system to compute the best sequence of inputs over a prediction horizon.
The first input from this planned sequence is then applied.
Since MPC is designed for linear systems, it has been used with linear models, such as the LIP model by Hamed et al.~\cite{hamed2020quadrupedal} or first-order Taylor approximation of the SRBD model by different authors~\cite{ding2021representation,ding2019real,risbourg2022real,dicarlo2018dynamic,chi2022linearization,farshidian2017real}.
Since linearization limits the applicability of MPC, some works used a non-linear MPC (NMPC)~\cite{norby2022quad,rathod2021model}.

Whole-body control (WBC) addresses the problem by defining a series of control tasks on the full dynamic model.
Such tasks, each defined with different priority, are then hierarchically solved by third-party quadratic programming (QP) solvers~\cite{risbourg2022real,raiola2020simple,bellicoso2017dynamic,li2023real,hutter2014quadrupedal}.
Sometimes, WBC is used in conjunction with other controllers that generate reference trajectories for the body~\cite{risbourg2022real,li2023real}.
	
Both WBC and MPC are based on optimization techniques that require solving computationally intensive non-linear optimization problems and may encounter convergence issues~\cite{li2023real}.
For instance, for NMPC it is often challenging to meet real-time requirements because the solver can get stuck in local minima~\cite{fahmi2019passive}.
Moreover, in a numerical optimization solver algorithm, each inequality constraint can activate a distinct operational branch, making the temporal evolution of the algorithm difficult to predict.
Usually, these optimization algorithms rely on third-party libraries~\cite{raiola2020simple,bellicoso2017dynamic,rathod2021model,dicarlo2018dynamic,farshidian2017robust} that may not be able to guarantee predictable temporal behavior, due to dynamic allocation or non-optimized memory usage.
These factors are particularly undesirable in hard real-time systems that must guarantee a result within predefined time bounds.

Model-free control methods encompass several techniques, that do not depend on the choice of a model.
The two most commonly used approaches of this type are described below.

Central pattern generator (CPG) is a biomimetic control method that involves constructing multiple periodic oscillators to generate rhythmic motion by coordinating their connections.
Since parameter tuning can be complex, this technique is often combined with genetic algorithms or reinforcement learning to optimize the parameters~\cite{zhang2023synloco}.

Reinforcement Learning (RL) is a learning technique that enables the robot to learn to walk based on its experience, by choosing an action from a set of possible actions it can perform in a given state and receiving feedback on the quality of the chosen action in terms of reward. 
Since the sets of possible states and actions for a quadruped robot are very large, it is necessary to use neural networks to represent value functions and policies, which could require too much memory~\cite{Gogineni2024SwiftRL:} and computational power~\cite{Sahoo2021MemOReL:,Li2019Implementation,Baranwal2021ReLAccS:} for a generic embedded system, that is, without a dedicated hardware accelerator based on FPGA~\cite{Sahoo2021MemOReL:}.
For example, in the framework presented by Tan et al.~\cite{tan2021hierarchical}, the RL component is executed on a dedicated server rather than on the robot itself, since the robot's onboard computer lacked sufficient computational power to run RL online. 
Therefore, although RL is a very powerful tool for high-level tasks that are difficult to model mathematically, we chose not to integrate it in our framework, which focuses on low-level tasks that are easily modeled, in order to be predictably and effortlessly executed on embedded systems with hard real-time constraints and limited computational resources.

\subsection{Feet Management}

This task consists of finding the alternation of foot movements that allows the robot to walk and run in the environment.
Since it is a complex problem, it is handled by dividing it into various subproblems described below.

\subsubsection{Gait planning} \label{sssec:soagait}

The gait is the pattern of limb movements during locomotion.
Kajita and Espiau~\cite{kajita2008legged} provided definitions commonly adopted for characterizing the gait.

Since the gait is repeated periodically, it is defined as a cycle that lasts for a given period $T$.
Being alternately lifted and lowered to walk, each foot goes through two phases during the gait cycle: 
a support phase, during which the foot is on the ground, 
and a swing phase, during which the foot is lifted. 
Assuming that each foot remains lifted for the same amount of time as the others, a duty factor $\beta \in \quadre{0;1}$ can be defined such that the support phase lasts $T_{st} = \beta T$ and the swing phase lasts $T_{sw} = \tonde{1-\beta} T$ for each foot.

Additionally, since feet cannot all be lifted simultaneously, they are lifted at different times, appropriately phased with each other depending on the type of gait.
Consequently, each foot $i$ leaves the ground at the instant $t_{0,i}$ (\textit{lift-off time}) and touches the ground again at the instant $t_{0,i} + T_{sw}$ (\textit{touch-down time}).
In our work, we defined a policy for choosing the parameters $T$, $T_{sw}$, and $\beta$ based on the input commands to the framework and the constraints on the footholds, as explained in Section~\ref{sssec:period}.

A common convention is to number the feet on the left with odd numbers and the feet on the right with even numbers, starting from the front feet.
Therefore, in a quadruped robot, foot 1 is the front-left (FL) foot, foot 2 is the front-right (FR) foot, foot 3 is the rear-left (RL) foot, and foot 4 is the rear-right (RR) foot.

Defining the function $\gamma : \graffe{1:4} \to \graffe{1:4}$ such that $i = \gamma\di{j}$ is the foot to be lifted as $j$-th, the gait that has the maximum stability in a certain direction of motion is called the \textit{crawl gait} and is characterized by the lifting sequences reported in Table~\ref{tab:crawlgait}.

\begin{table}[h]
	\centering
	\begin{tabular}{|c|c|c|c|c|}
		\hline \textbf{Motion} & $\gamma\di{1}$ & $\gamma\di{2}$ & $\gamma\di{3}$ & $\gamma\di{4}$ \\
		\hline Forwards & FL & RR & FR & RL \\
		\hline Backwards & FL & RL & FR & RR \\
		\hline Leftwards & FL & RR & RL & FR \\
		\hline Rightwards & FL & FR & RL & RR \\
		\hline Counterclockwise & FL & RL & RR & FR \\
		\hline Clockwise & FL & FR & RR & RL \\
		\hline
	\end{tabular} 
	\caption{Crawl gait according to motion.}
	\label{tab:crawlgait}
\end{table}

\begin{figure}[h]
	\includegraphics[width=\linewidth]{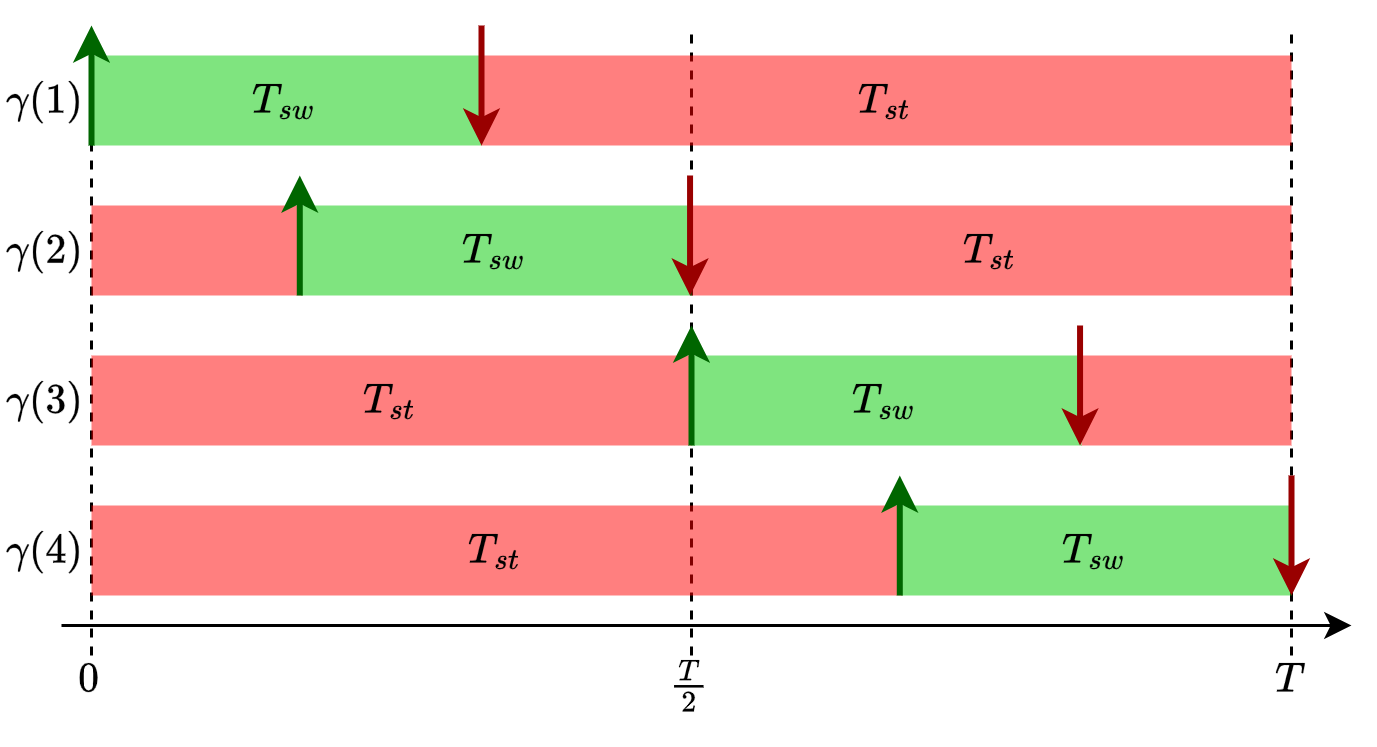}
	\caption{Gait graph of a wave gait, where dark green and dark red arrows indicate the lift-off and touch-down times respectively, with respect to the period $T$ and the swing time $T_{sw}$.}
	\label{fig:gaitgraph}
\end{figure}

The crawl gait that maximizes the longitudinal stability margin is called the \textit{wave gait} and is characterized by the lift-off times pattern in the table of Equation~\eqref{eq:liftofftimes}.
\begin{equation} \label{eq:liftofftimes}
	\begin{array}{|c|c|c|c|c|}
		\hline i & \gamma\di{1} & \gamma\di{2} & \gamma\di{3} & \gamma\di{4} \\
		\hline \vphantom{\dfrac{T}{2}} t_{0,i} & 0 & \frac{T}{2} - T_{sw} & \frac{T}{2} & T - T_{sw} \\
		\hline
	\end{array}
\end{equation}
Such lift-off times can be visualized in the gait graph shown in Figure~\ref{fig:gaitgraph}.

In our work, we adapted the state-of-the-art convention reported above by defining a policy that handles cases where the input commands to the framework do not correspond to a motion in a single direction, as explained in Section~\ref{sssec:gaitsch}.

\subsubsection{Foothold planning}

Foothold planning is the problem of choosing where to place the foot after it has been lifted to take a step.
Terrains without holes or deformations allow much more freedom in foot landing choice.
Many works~\cite{norby2022quad, risbourg2022real, raiola2020simple, li2023real, rathod2021model} have chosen to refer to the Raibert heuristic defined for a one-legged robot by Raibert~\cite{raibert1986legged}, introducing variations or corrections.

This heuristic approach was avoided in this work, instead adopting an algorithm based on the concept of imaginary wheels, explained in Section~\ref{sssec:wheelcmds}.

\subsubsection{Foot trajectories} \label{sssec:soatraj}

Each foot must follow a trajectory that lifts it and places it in the desired foothold.
Therefore, after computing the footholds, it is necessary to connect them with a proper trajectory.
Several trajectories have been proposed in various works.

Tan et al.~\cite{tan2021hierarchical} used polynomial trajectories whose parameters are calculated by a neural network.
Barasoul et al.~\cite{barasuol2013reactive} used elliptical trajectories inspired by the CPG, which are cut when the foot touches the ground.
Li et al.~\cite{li2023real} used second-order Bézier curves to interpolate the initial and final footholds by raising the foot to a certain height.
Raiola et al.~\cite{raiola2020simple} used a parametric semi-elliptical trajectory in a local foot frame.
Shi et al.~\cite{shi2021structural} used a cycloidal trajectory (described by a point on the border of a rolling wheel) in a global frame.

All these solutions have a common inconsistency: despite the control techniques used for the body are based on the assumption that the feet have both zero velocity and acceleration when on the ground, they do not take this aspect into account when choosing foot trajectories.
However, in the proposed framework, this assumption is taken into account, as explained in Section~\ref{sssec:trajgen}.

%

	\section{System model} \label{sec:model}

This section describes the quadruped robot model on which \NAME\ was built. Generally, a quadruped robot consists of a floating base, named \textit{body}, to which four serial kinematic chains, named \textit{legs}, are connected.
Each leg comprises three revolute joints, and its end-effector is named \textit{foot}.
The quadruped robot model consists of a kinematic part, detailed later in Section~\ref{ssec:kinematics}, and a dynamic part, detailed later in Section~\ref{ssec:dynamics}.

For the comprehension of the reader, here is a summary of the notation used throughout this article:
small capital letters indicate points;
over right arrows indicate vectors in space;
framed capital letters indicate reference systems (frames);
underlines generally indicate n-tuples;
underlines with a superscript lowercase letter or number indicate the coordinates of the vectors in the frame specified by the superscript;
$R_a^b$ is a rotation matrix from a frame $\SdR{A}$ to a frame $\SdR{B}$;
$\quadre{\ua \times}$ is the skew-symmetric matrix of the vector $\ua \in \bR^3$ such that $\quadre{\ua \times}\ub = \ua \times \ub \forevery \ub \in \bR^3$;
and $\graffe{\ui, \uj, \uk}$ is the canonical basis of $\bR^3$.

\subsection{Kinematics} \label{ssec:kinematics}

\begin{figure}[h]
	\includegraphics[width=\linewidth]{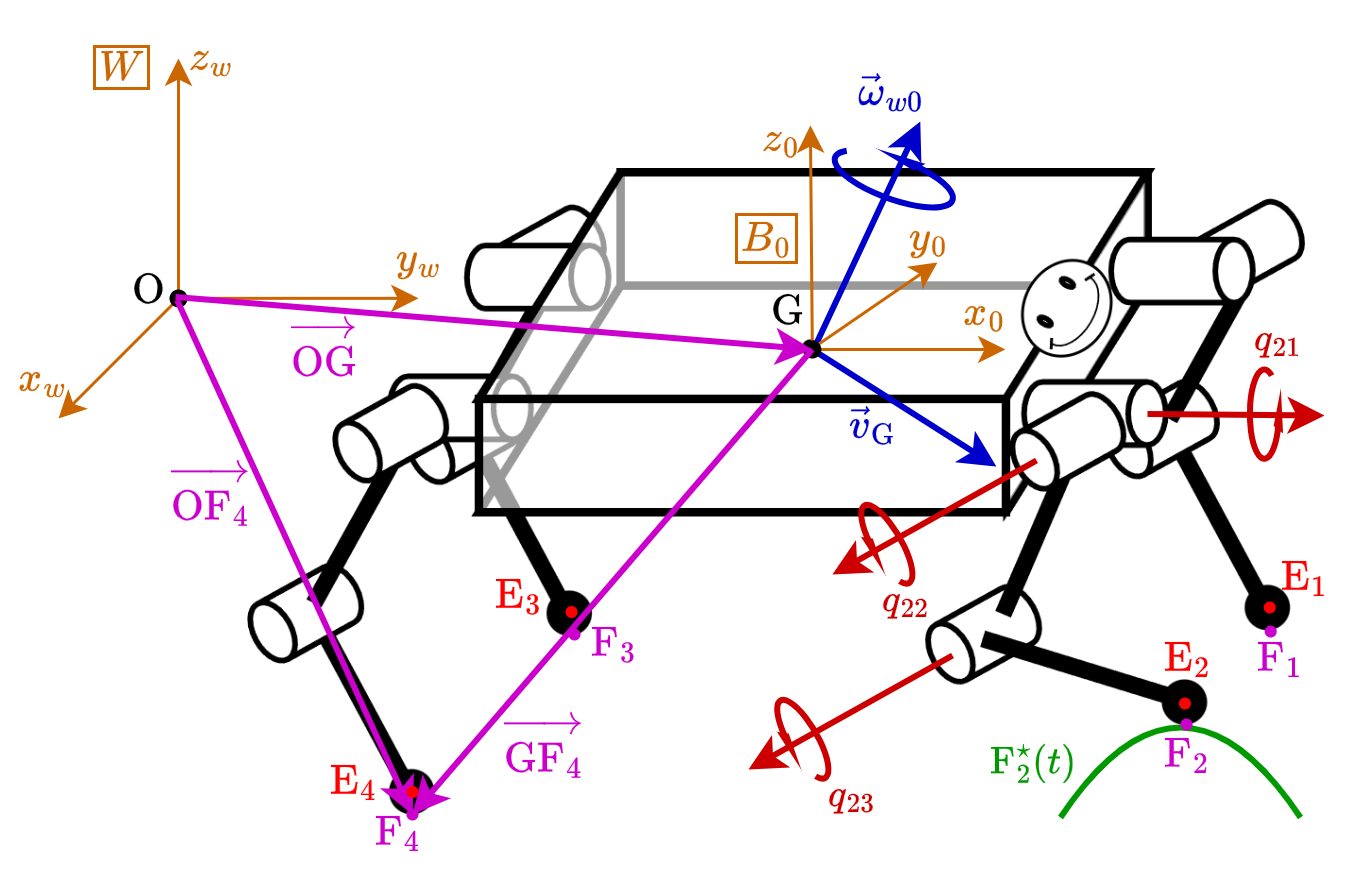}
	\caption{The orange arrows are the \textit{world} and \textit{body} frames.
	The purple arrows are the position vectors of the center of mass and foot 4.
	The blue arrows are linear and angular velocities of the body.
	The red arrows are the joint angles of leg 2.
	The red points are the feet centers $\pE_i$.
	The purple points are the feet contact points $\pF_i$.
	The green line is the trajectory followed by foot contact point $\pF_2$ lifted to make a step.}
	\label{fig:kinematics}
\end{figure}

This subsection defines the kinematic quantities necessary to describe the motion of the robot.
Such kinematic quantities are all illustrated in Figure~\ref{fig:kinematics}.

To correctly model the quadruped robot, the following frames are defined a fixed frame $\SdR{W}$ named \textit{World}, with origin $\pO$, such that $\ug^w = -g\uk$, and a frame $\SdR{B_0}$ named \textit{Body} attached to the body, with origin in the center of mass $\pG$ of the body.

The center of mass $\pG$ can translate freely, then it has three degrees of freedom, and its coordinates are $\uvec{\pO\pG}^w = \matrice{x & y & z}^\top$.
The body can rotate freely, so its orientation $R_0^w\di{\uphi} = R_z\di{\yaw} R_y\di{\pitch} R_x\di{\roll}$ has three degrees of freedom $\uphi = \matrice{\roll & \pitch & \yaw}^\top$, corresponding to roll $\roll$, pitch $\pitch$, and yaw $\yaw$.
Therefore, the body has six degrees of freedom, represented by $\uq_0 = \matrice{\uvec{\pO\pG}^w \\ \uphi} \in \bR^6$.
In addition, each leg $i$ is made up of three revolute joints, each of which rotates by an angle $q_{ij}$, therefore it has three degrees of freedom represented by $\uq_i = \matrice{q_{i1} & q_{i2} & q_{i3}}^\top$.
Since the robot has four legs, the robot has a total of 18 degrees of freedom.

The coordinates in $\SdR{B_0}$ of the linear velocity of the center of mass $\vv_\pG$ are given by $\uv_\pG^0 = \quadre{R_0^w\di{\uphi}}^\top \duvec{\pO\pG}^w$,
and those of the angular velocity of the body $\vom_{w0}$ are given by
$\uom_{w0}^0 = \dot{\roll} \ui + \dot{\pitch} R_x^\top\di{\roll} \uj + \dot{\yaw} R_x^\top\di{\roll} R_y^\top\di{\pitch} \uk = \Omega_{w0}^0\di{\uphi} \duphi$.
Together, they form the twist, whose coordinates in $\SdR{B_0}$ are given by
\begin{equation} \label{eq:twist}
	\uV_{w0}^0 = \matrice{\uv_\pG^0 \\ \uom_{w0}^0} = \matrice{\quadre{R_0^w\di{\uphi}}^\top \duvec{\pO\pG}^w \\ \Omega_{w0}^0\di{\uphi} \duphi}
	= J_{w0}^0\di{\uphi} \duq_0.
\end{equation}

Each foot $i$ is assumed to be a sphere with center $\pE_i$ and radius $\zeta$, that touches the ground at the contact point $\pF_i$.
Therefore, the center $\pE_i$ has coordinates
\begin{equation} \label{eq:GE}
	\uvec{\pO\pE_i}^w\di{\uq_0, \uq_i} = \uvec{\pO\pG}^w + R_0^w\di{\uphi} \uvec{\pG\pE_i}^0\di{\uq_i},
\end{equation}
where $\uvec{\pG\pE_i}^0\di{\uq_i}$ depends on the chosen robot, and the contact point $\pF_i$ has coordinates
\begin{equation} \label{eq:OF}
	\uvec{\pO\pF_i}^w = \uvec{\pO\pE_i}^w - \zeta \uk = \uvec{\pO\pG}^w + R_0^w \uvec{\pG\pF_i}^0.
\end{equation}
By differentiating Equation~\eqref{eq:OF}, we obtain
\begin{equation} \label{eq:dOF}
	\duvec{\pO\pF_i}^w = R_0^w \tonde{A_i \uV_{w0}^0 + \duvec{\pG\pF_i}^0},
\end{equation}
where we have set
\begin{equation} \label{eq:A_i}
	A_i \est \matrice{I_3 & - \quadre{\uvec{\pG\pF_i}^0 \times}}.
\end{equation}
By differentiating Equation~\eqref{eq:dOF}, we obtain
\begin{equation} \label{eq:ddOF}
	\dduvec{\pO\pF_i}^w = R_0^w \tonde{\uh_i + A_i \duV_{w0}^0 + \dduvec{\pG\pF_i}^0},
\end{equation}
where we have defined
\begin{equation} \label{eq:uh_i}
	\uh_i \est \uom_{w0}^0 \times \tonde{A_i \uV_{w0}^0 + 2 \duvec{\pG\pF_i}^0}.
\end{equation}
From Equation~\eqref{eq:OF} we derive
\begin{equation} \label{eq:GF}
	\uvec{\pG\pF_i}^0 = \uvec{\pG\pE_i}^0 - \zeta \quadre{R_0^w}^\top \uk.
\end{equation}
By differentiating Equation~\eqref{eq:GF}, we obtain
\begin{equation} \label{eq:dGF}
	\duvec{\pG\pF_i}^0 = J_{\pG\pE_i}^0\di{\uq_i} \duq_i + \uom_{w0}^0 \times \zeta \quadre{R_0^w}^\top \uk,
\end{equation}
where we have defined $J_{\pG\pE_i}^0\di{\uq_i} \est \diff{\uvec{\pG\pE_i}^0}{\uq_i}$.

The moment that foot $i$ touches the ground, its position $\uvec{\pO\pF_i}^w$ stays constant, then $\duvec{\pO\pF_i}^w = \uoh$ and $\dduvec{\pO\pF_i}^w = \uoh$.
If we define
\begin{equation}
	\sigma_i \est 
	\begin{cases*}
		1 & if foot $i$ is grounded \\ 
		0 & if foot $i$ is not grounded
	\end{cases*}
\end{equation}
for each foot $i$, it always holds $\sigma_i \duvec{\pO\pF_i}^w = \uoh$ and $\sigma_i \dduvec{\pO\pF_i}^w = \uoh$,
thus, for Equations~\eqref{eq:dOF} and \eqref{eq:ddOF}, we have
\begin{equation} \label{eq:grf1}
	\sigma_i \tonde{A_i \uV_{w0}^0 + \duvec{\pG\pF_i}^0} = \uoh
\end{equation}
and
\begin{equation} \label{eq:grf2}
	\sigma_i \tonde{\uh_i + A_i \duV_{w0}^0 + \dduvec{\pG\pF_i}^0} = \uoh.
\end{equation}

\subsubsection{Uncontrollable motions} \label{sssec:uncm}

Let's analyze how the relative motion of the feet on the ground influences the motion of the body: 
by applying Equation~\eqref{eq:grf1} to all the feet, we obtain
\begin{equation} \label{eq:unc1}
	A \uV_{w0}^0 + \Sigma \duvec{\pG\pF}_{1:4}^0 = \uoh,
\end{equation}
where we defined
\begin{equation} \label{eq:matrixA}
	A \est \matrice{\sigma_1 A_1^\top & \sigma_2 A_2^\top & \sigma_3 A_3^\top & \sigma_4 A_4^\top}^\top,
\end{equation}
\begin{equation}
	\duvec{\pG\pF}_{1:4}^0 \est \matrice{\tonde{\duvec{\pG\pF_1}^0}^\top & \tonde{\duvec{\pG\pF_2}^0}^\top & \tonde{\duvec{\pG\pF_3}^0}^\top & \tonde{\duvec{\pG\pF_4}^0}^\top }^\top,
\end{equation}
and
\begin{equation}
	\Sigma \est \diag{\sigma_1 I_3, \sigma_2 I_3, \sigma_3 I_3, \sigma_4 I_3}.
\end{equation}
Equation~\eqref{eq:unc1} represents a system of 12 equations with 6 variables, because we want to find $\uV_{w0}^0 \in \bR^6$ as a function of $\duvec{\pG\pF}_{1:4}^0 \in \bR^{12}$.
Equation~\eqref{eq:unc1} does not have a solution if $ \uV_{w0}^0 \in \cN\op{A} $, therefore the null space $\cN\op{A}$ can be considered the space of uncontrollable motions.
If $r$ denotes the rank of matrix $A$, then the body has only $r \le 6$ controllable degrees of freedom, because only $r$ equations are actually linearly independent.
Since matrix $A$ changes depending on the contact flags $\sigma_{1:4}$, 
the null space $\cN\op{A}$ has been calculated for each possible number of grounded feet $N$, thus deriving the corresponding value of its rank as $r = 6 - \dim \cN\op{A}$.
For brevity, the computation procedure is not reported here, but the resulting values of $r$ for each value of $N$ are shown in Table~\ref{tab:r}.
Knowing a priori the values of $r$ as a function of $N$ will allow us to make the LDQ decomposition robust to numerical approximations, as explained later in Section~\ref{sssec:ldq}.

\begin{table}[h]
	\centering
	\begin{tabular}{|c|c|c|c|c|c|}
		\hline
		$N$ & 0 & 1 & 2 & 3 & 4 \\
		\hline
		$r$ & 0 & 3 & 5 & 6 & 6 \\
		\hline
	\end{tabular}
	\caption{Values of $r=\rank A$ corresponding to the number of grounded feet $N$ computed offline.}
    \label{tab:r}
\end{table}
\subsection{Dynamics} \label{ssec:dynamics}

\begin{figure}[h]
	\includegraphics[width=\linewidth]{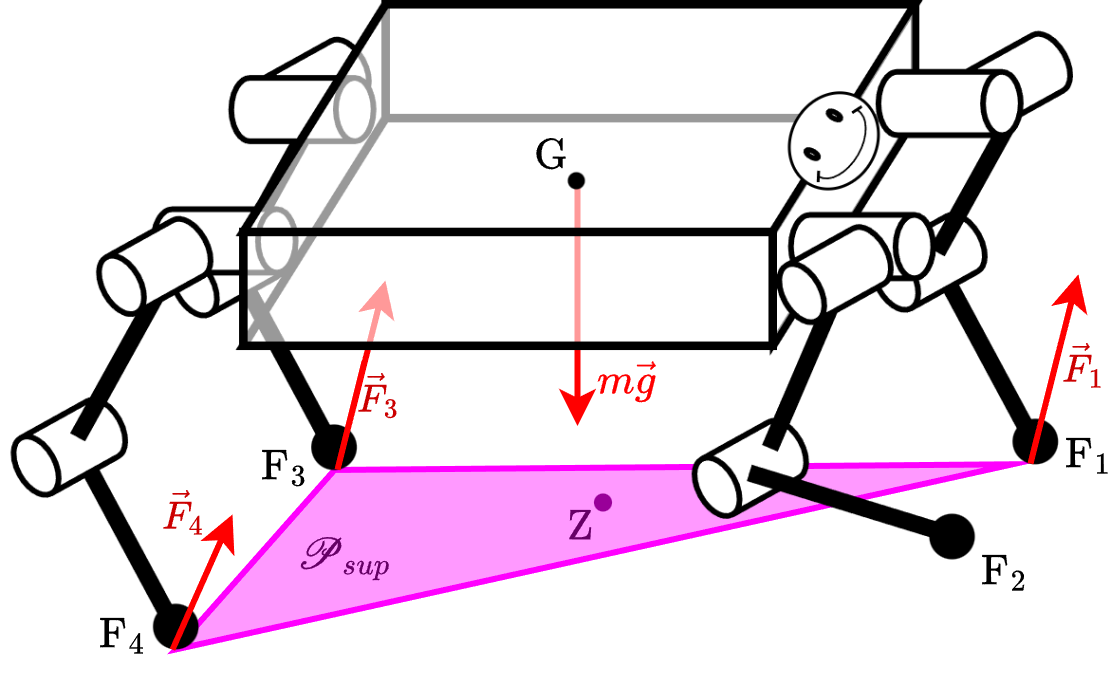}
	\caption{The red arrows are the external forces acting on the robot. The magenta polygon is the Support Polygon. The magenta point $\pZ$ is the Zero Moment Point.}
	\label{fig:dynamics}
\end{figure}

This subsection defines the dynamic quantities necessary to describe the causes of the motion of the robot.
Such dynamic quantities are all illustrated in Figure~\ref{fig:dynamics}.

The weight force $\uF_g^w = m \ug^w = - m g \uk$ acts on the center of mass $\pG$.
A contact force $\uF_{\perp,i}^w = F_{\perp,i}^w \uk$ such that $F_{\perp,i}^w \ge 0$, perpendicular to the ground, and a friction force $\uF_{\parallel,i}^w$ such that $\uk^\top \uF_{\parallel,i}^w = 0$ and $\bignorm{\uF_{\parallel,i}^w} \le \mu_s F_{\perp,i}^w$, parallel to the ground, act on the contact point $\pF_i$ of each foot $i$.
For simplicity, in this work we assume $\mu_s \simeq \oo$.
We define $\uF_i^w \est \uF_{\perp,i}^w + \uF_{\parallel,i}^w$ as the resulting force on each contact point $\pF_i$.
Since the forces on foot $i$ exist only if it touches the ground, it holds $\uF_i^w = \sigma_i \uF_i^w$.

According to the Newton-Euler equations of dynamics, the dynamic equilibrium of the forces acting on the quadruped robot is given by
\begin{equation} \label{eq:dyn}
	M_\pG^0 \duV_{w0}^0 + \quadre{\ads \uV_{w0}^0} M_\pG^0 \uV_{w0}^0 = \matrice{\uF_g^0 \\ \uoh} + \sum_{i=1}^4 \sigma_i A_i^\top \uF_i^0,
\end{equation}
where $M_\pG^0 \est \diag{mI_3, \I_\pG^0}$ is the generalized inertia tensor of the body, with $m$ being the mass and $\I_\pG^0$ the inertia tensor of the body, and
\begin{equation} \label{eq:ads}
	\ads \uV_{w0}^0 = \matrice{\quadre{\uom_{w0}^0 \times} & O \\ \quadre{\uv_\pG^0 \times} & \quadre{\uom_{w0}^0 \times}}.
\end{equation}

\subsubsection{Zero Moment Point (ZMP)} \label{sssec:zmp}

As reported by Kajita and Espiau~\cite{kajita2008legged}, the convex hull of the contact points of the feet that touch the ground is named \textit{Support Polygon}, and it can be defined mathematically by
\begin{equation} \label{eq:P_sup}
	\sP_{sup} = \graffe{\uvec{\pO\pP}^w = \sum_{i=1}^{4} \lambda_i \sigma_i \uvec{\pO\pF_i}^w \tc \sum_{i=1}^{4} \lambda_i = 1 , \lambda_i \ge 0}.
\end{equation}
Defining the inertial wrench as
\begin{equation}
	\uW_V^0 \est M_\pG^0 \duV_{w0}^0 + \quadre{\ads \uV_{w0}^0} M_\pG^0 \uV_{w0}^0,
\end{equation}
Equation~\eqref{eq:dyn} can be rewritten as
\begin{equation} \label{eq:dyn2}
	\uW_V^0 = \matrice{\uF_V^0 \\ \uM_V^0} = \matrice{\uF_g^0 \\ \uoh} + \sum_{i=1}^4 \sigma_i A_i^\top \uF_i^0,
\end{equation}
where the inertial wrench $\uW_V^0$ is decomposed into its components, i.e., the inertial force $\uF_V^0$ and the inertial moment $\uM_V^0$.
Starting from Equation~\eqref{eq:dyn2}, it can be proved~\cite{kajita2008legged} that there exists a point $\pZ$ named \textit{Zero Moment Point} (ZMP) such that
\begin{equation} \label{eq:zmp9}
	\uvec{\pO\pZ}^w \est \uvec{\pO\pG}^w - \frac{z \uF_{Vg}^w - \uk \times \uM_V^w}{\uk^\top \uF_{Vg}^w} \in \sP_{sup},
\end{equation}
where $\uF_{Vg}^w = R_0^w \tonde{\uF_V^0 - \uF_g^0}$ and $\uM_V^w = R_0^w \uM_V^0$.
In the case where $\uvec{\pO\pZ}^w \notin \sP_{sup}$, the dynamic equilibrium would not be maintained and the robot would become unstable. Therefore, the stability of the robot is guaranteed as long as $\uvec{\pO\pZ}^w \in \sP_{sup}$.

	\section{Proposed framework} \label{sec:framework}

This section presents in detail the components of \NAME, by describing its internal sub-modules.

As shown in Figure~\ref{fig:fwbd}, the framework receives high-level input commands, which can be sent either by a user through a joystick or by a high-level planner, and produces the joint velocities $\duq_{1:4}^\star$ to move the quadruped robot as required. 
The robot feeds back to the framework the actual joint positions $\uq_{1:4}$ and the contact signals $\sigma_{1:4}$.

Internally, the framework is divided into three main blocks: the \textit{Feet Manager}, the \textit{Body Reference}, and the \textit{Body Manager}, which are described below.

\subsection{Feet Manager} \label{ssec:feetman}

\begin{figure}[h]
	\centering%
	\includegraphics[width=\linewidth]{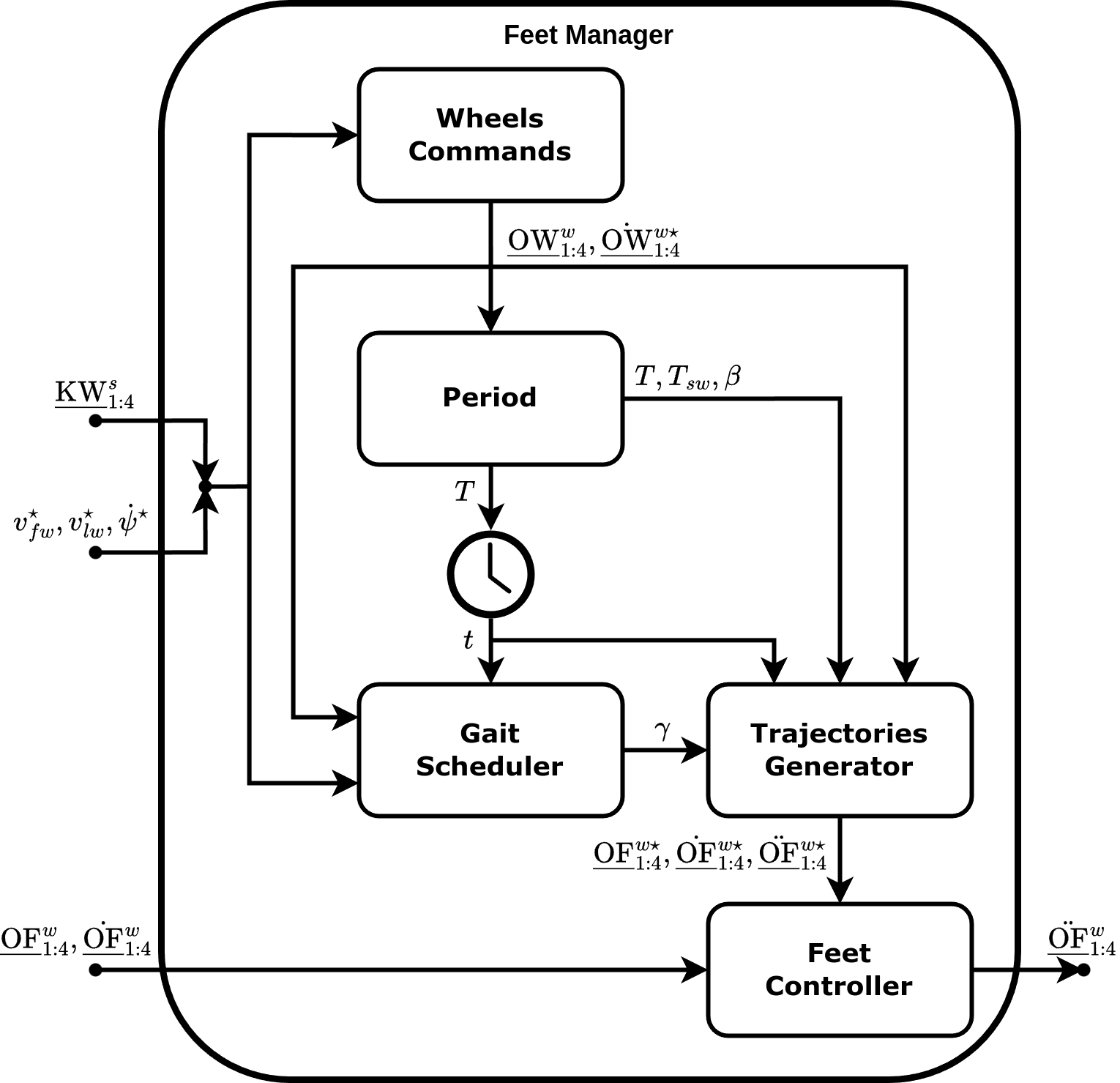}%
	\caption{Feet manager block diagram.}%
	\label{fig:feetbd}
\end{figure}

This module receives high-level velocity commands (longitudinal velocity $v_{fw}^\star$, lateral velocity $v_{lw}^\star$, and yaw rate $\dot{\yaw}^\star$) and converts them into foot acceleration commands $\dduvec{\pO\pF}_{1:4}^w$, forwarded to the \textit{Body Manager}.
The block diagram of the \textit{Feet Manager} is illustrated in Figure~\ref{fig:feetbd} and consists of six modules described in the following subsubsections.

\subsubsection{Wheels Commands} \label{sssec:wheelcmds}

\begin{figure}[h]
	\includegraphics[width=\linewidth]{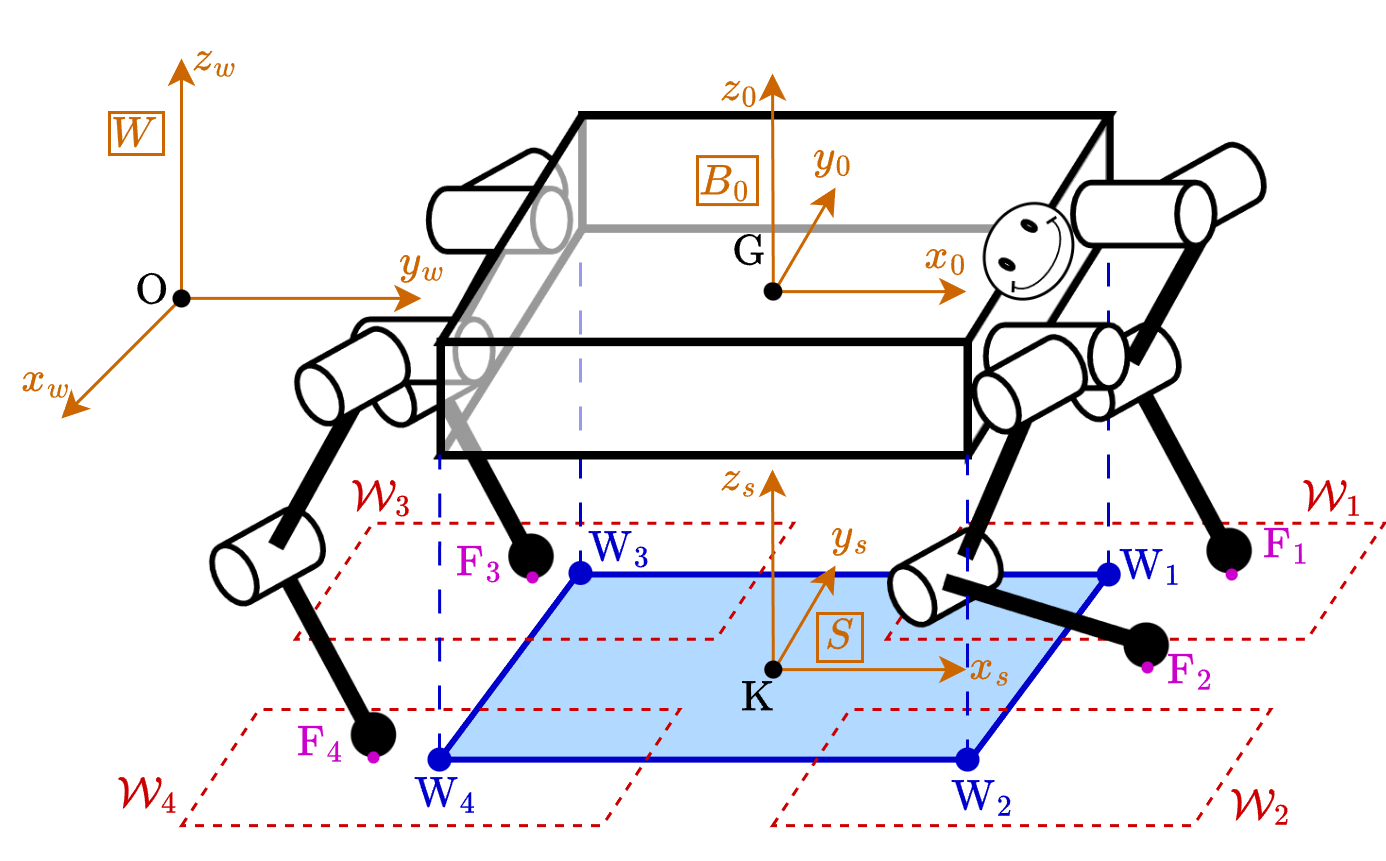}
	\caption{Shadow of the body (blue rectangle) delimited by imaginary wheels (blue dots $\pW_i$). The \textit{world}, \textit{shadow} and \textit{body} frames are denoted in orange. The red dotted rectangles denote the feet workspaces $\cW_i$.}
	\label{fig:ws}
\end{figure}

To compute footholds, we assume that the projection of the body on the ground (named \textit{shadow}) has 4 imaginary wheels at points $\pW_i \forevery i \in \graffe{1:4}$, as shown in Figure~\ref{fig:ws}.

This block transforms $v_{fw}^\star$, $v_{lw}^\star$, and $\dot{\yaw}^\star$ into positions $\uvec{\pO\pW}_{1:4}^w$ and desired velocities $\duvec{\pO\pW}_{1:4}^{w\star}$ in the $\SdR{W}$ frame of such imaginary wheels.
This data will then be used by the \textit{Period} block (see Section~\ref{sssec:period}), the \textit{Gait Scheduler} block (see Section~\ref{sssec:gaitsch}), and the \textit{Trajectories Generator} block (see Section~\ref{sssec:trajgen}).

For convenience, an intermediate frame $\SdR{S}$ (\textit{shadow}) is attached to the projection of the body on the ground (assumed flat), as shown in Figure~\ref{fig:ws}.
Therefore, for the origin $\pK$ of $\SdR{S}$ we have $\uvec{\pO\pK}^w = x \ui + y \uj$,
$R_s^w = R_z\di{\yaw}$,
$\uvec{\pK\pG}^s = z \uk$,
$R_0^s = R_y\di{\pitch} R_x\di{\roll}$,
and its commanded velocity in $\SdR{W}$ is $\duvec{\pO\pK}_\star^w = R_z\di{\yaw} \bigtonde{v_{fw}^\star \ui + v_{lw}^\star \uj}$. Hence, the position of wheel $i$ is given by
\begin{equation} \label{eq:OW}
	\uvec{\pO\pW_i}^w = \uvec{\pO\pK}^w + R_s^w \uvec{\pK\pW_i}^s = x \ui + y \uj + R_z\di{\yaw} \uvec{\pK\pW_i}^s
\end{equation}
and its commanded velocity is given by
\begin{multline} \label{eq:dOW}
	\duvec{\pO\pW_i}_\star^w = \duvec{\pO\pK}^w + \dot{R}_s^w \uvec{\pK\pW_i}^s =
	\\= R_z\di{\yaw} \tonde{v_{fw}^\star \ui + v_{lw}^\star \uj + \dot{\yaw}^\star \uk \times \uvec{\pK\pW_i}^s},
\end{multline}
where $\uvec{\pK\pW_i}^s$ is provided by the \textit{Body Manager} as
\begin{equation}
	\uvec{\pK\pW_i}^s = \matrice{\ui & \uj & \uoh}^\top R_y\di{\pitch} R_x\di{\roll} \uvec{\pG\pW_i}^0
\end{equation}
because it depends on current roll and pitch angles and the internal parameter $\uvec{\pG\pW_i}^0$.

Since the feet are required to move as much as the imaginary wheels during a period, their positions and desired velocities are used to compute the initial and final footholds:
it is desired that foot $i$ starts from the initial position $\uvec{\pO\pW_i}^w\di{0}$ and reaches the final position $\uvec{\pO\pW_i}^w\di{T}$ of wheel $i$.
After updating the values of $\uvec{\pO\pW_i}^w$ and $\duvec{\pO\pW_i}_\star^w$ online using Equations~\eqref{eq:OW} and \eqref{eq:dOW}, the initial foothold is obtained as
\begin{equation} \label{eq:footstart}
	\uvec{\pO\pW_i}^w\di{0} = \uvec{\pO\pW_i}^w - t \duvec{\pO\pW_i}_\star^w
\end{equation}
and the final foothold as
\begin{equation} \label{eq:footend}
	\uvec{\pO\pW_i}^w\di{T} = \uvec{\pO\pW_i}^w + \tonde{T-t} \duvec{\pO\pW_i}_\star^w,
\end{equation}
where $t$ is the current time step $t$ marked by the clock (see Section~\ref{sssec:clock}), and it is assumed that the desired velocity of the wheel remains constant throughout the gait period $T$.
This assumption holds as long as the high-level velocity commands remain constant, but it has been experimentally proven effective also when the commands change, thanks to the fact that $\uvec{\pO\pW_i}^w$, $\duvec{\pO\pW_i}_\star^w$, and $T$ are updated at each execution of \NAME.

\subsubsection{Period} \label{sssec:period}

This block computes the gait period $T$, the duty factor $\beta$, and the swing time $T_{sw}$ (see Section~\ref{sssec:soagait} for definitions) as a function of the current positions and desired velocities of the imaginary wheels received from the \textit{Wheels Commands} block.

In order for the step to be physically feasible, the foot position must be contained within the \textit{workspace} region (shown in Figure~\ref{fig:ws}), defined for each foot $i$ as
\begin{multline} \label{eq:ws}
	\cW_i \est \graffe{\uvec{\pO\pP}^w = \uvec{\pO\pK}^w + R_s^w \uvec{\pK\pP}^s \tc
	\newlineparentheses \ui^\top \uvec{\pK\pP}^s \in \bigquadre{x_{i,min}^s; x_{i,max}^s}, \uj^\top \uvec{\pK\pP}^s \in \bigquadre{y_{i,min}^s; y_{i,max}^s}}
\end{multline}
such that $\uvec{\pO\pW_i}^w\di{\tau} \in \cW_i \forevery \tau \in \quadre{0;T}$.
Currently, the workspaces are defined to ensure that the feet do not cross each other during the gait, but they can be modified to handle other situations.

For a feasible step, it must hold $T_{sw} \in \quadre{T_{sw}^{min}; T_{sw}^{max}}$ and $\beta \in \quadre{\beta_{min}; \beta_{max}} \subset \quadre{0;1}$.
As a consequence, it must also hold $T \in \quadre{T_{min}; T_{mid}} \cup \quadre{T_{mid}; T_{max}}$, where $T_{min} = \frac{T_{sw}^{min}}{1 - \beta_{min}}$, $T_{mid} =  \frac{T_{sw}^{min}}{1 - \beta_{max}}$, and $T_{max} =  \frac{T_{sw}^{max}}{1 - \beta_{max}}$.

\vspace{2mm}
For a given period $T$, a stable gait can be achieved by maximizing the support time $T_{st} = \beta T = T - T_{sw}$, which means maximizing $\beta$ and minimizing $T_{sw}$.
On the other hand, an efficient gait (i.e., with the fewest steps) is achieved by maximizing the period $T$.
Since for Equation~\eqref{eq:footend} it must hold that
\begin{equation} \label{eq:T}
	T = t + \frac{\norm{\uvec{\pO\pW_i}^w\di{T} - \uvec{\pO\pW_i}^w}}{\norm{\duvec{\pO\pW_i}_\star^w}} \forevery i \in \graffe{1:4},
\end{equation}
the maximum period $T$ that satisfies $\uvec{\pO\pW_i}^w\di{T} \in \cW_i$ is computed using the following procedure:
initialize $T$ with $T_{max}$ and, for each foot $i$, compute 
(i) the position $\uvec{\pO\pW_i}^w\di{T}$; 
(ii) the position $\uvec{\pO\pW_i}_{\in}^w\di{T_i} \in \cW_i$ closest to $\uvec{\pO\pW_i}^w\di{T}$, and 
(iii) the period candidate $T_i$ that would make the foot $i$ reach the boundaries of its workspace $\cW_i$, using Equation~\eqref{eq:T}.
Then, $T$ is set to the minimum of all $T_i$, and the maximum $\beta$ and the minimum $T_{sw}$ is computed to be compatible with their bounds.
Such a procedure is detailed in Algorithm~\ref{alg:T}.

\begin{algorithm}[h]
	\caption{Period Computation} 
	\label{alg:T}
	\begin{algorithmic}[1]
		\STATE $T \est T_{max}$
		\FOR{$i \in \graffe{1:4}$}
			\IF{$\norm{\duvec{\pO\pW_i}_\star^w} = 0$}
				\STATE $T_i \est T_{max}$
			\ELSE
				\STATE $\uvec{\pO\pW_i}^w\di{T} \est \uvec{\pO\pW_i}^w + \tonde{T-t} \duvec{\pO\pW_i}_\star^w$
				\STATE $\uvec{\pO\pW_i}_{\in}^w\di{T_i} \in \cW_i$ closest to $\uvec{\pO\pW_i}^w\di{T}$
				\STATE $T_i \est t + \frac{\norm{\uvec{\pO\pW_i}_{\in}
				^w\di{T_i} - \uvec{\pO\pW_i}^w}}{\norm{\duvec{\pO\pW_i}_\star^w}}$
			\ENDIF
		\ENDFOR
		\STATE $T \est \min \graffe{T_i \tc i \in \graffe{1:4}}$
		\IF{$T \in \quadre{T_{mid}; T_{max}}$}
			\STATE $\beta \est \beta_{max}$, $T_{sw} \est \tonde{1-\beta_{max}} T$
		\ELSIF{$T \in \quadre{T_{min}; T_{mid}}$}
			\STATE $\beta \est 1 - \frac{T_{sw}^{min}}{T}$, $T_{sw} \est T_{sw}^{min}$
		\ENDIF
	\end{algorithmic}
\end{algorithm}

\subsubsection{Clock} \label{sssec:clock}

This module provides the current time $t$, elapsed from the beginning of the current period.
It is incremented by the sampling time $T_s$ at each iteration and it is reset when $t \ge T$.

\subsubsection{Gait Scheduler} \label{sssec:gaitsch}

This block decides the sequence for lifting the feet according to the commands $v_{fw}^\star$, $v_{lw}^\star$, and $\dot{\yaw}^\star$, that is, it decides the gait schedule $\gamma$, as defined in Section~\ref{sssec:soagait}.
The schedule is then used by the \textit{Trajectories Generator} to compute the lift-off times according to Equation~\eqref{eq:liftofftimes} (see Section~\ref{sssec:trajgen}).

The gait schedules shown in Table~\ref{tab:crawlgait} are provided only for motions in a single direction, that is, for velocity commands where only one of $v_{fw}^\star$, $v_{lw}^\star$, and $\dot{\yaw}^\star$ is different from zero (see Table~\ref{tab:motcmd}).

\begin{table}[h]
	\centering
	\begin{tabular}{|c|c|c|c|}
		\hline \textbf{Motion} & $v_{fw}^\star$ & $v_{lw}^\star$ & $\dot{\yaw}^\star$ \\
		\hline Forwards & $> 0$ & 0 & 0 \\
		\hline Backwards & $< 0$ & 0 & 0 \\
		\hline Leftwards & 0 & $> 0$ & 0 \\
		\hline Rightwards & 0 & $< 0$ & 0 \\
		\hline Counterclockwise & 0 & 0 & $> 0$ \\
		\hline Clockwise & 0 & 0 & $< 0$ \\
		\hline
	\end{tabular} 
	\caption{Correspondence between motion and velocity commands.}
	\label{tab:motcmd}
\end{table}

Therefore, it is necessary to define a criterion that generalizes the choice of the gait schedule to all possible velocity commands.
The idea is to choose the gait schedule corresponding to the motion that is predominant compared to the others.

\begin{figure}[t]
	\includegraphics[width=\linewidth]{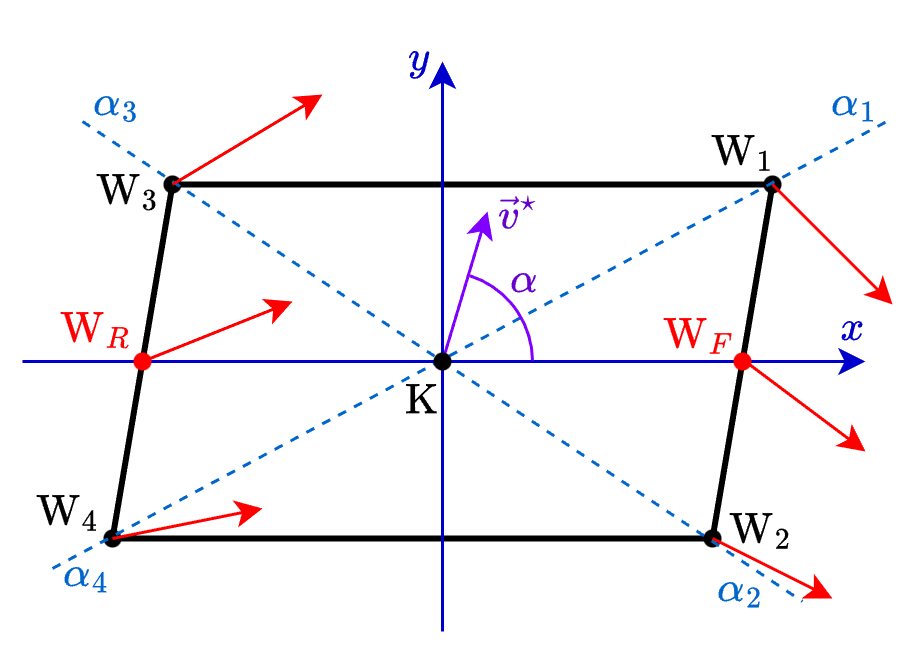}
	\caption{Illustration of the selection of the predominant motion: the red arrows are the velocities of the wheels, the purple angle $\alpha$ is formed by the commanded velocity $\vv^\star = v_{fw}^\star \vi_s + v_{lw}^\star \vj_s$ with the $x$ axis of the \textit{shadow} frame, the blue dashed lines split the sectors when gait scheduling changes.}
	\label{fig:predmot}
\end{figure}

Referring to Figure~\ref{fig:predmot}, the predominant motion is determined heuristically as follows.
The desired average velocity of the front imaginary wheels is compared with that of the rear wheels:
if the angle between them is larger than 90 degrees, then rotation is predominant over translation, and the gait schedule for clockwise or counterclockwise motion is selected.
When translation is predominant over rotation,
the half-lines that start from the origin and pass through the imaginary wheels divide the plane into four sectors: a front one, one on the right, one on the back, and one on the left.
The selected gait schedule is the one corresponding to that sector in which the commanded velocity vector lies.
This procedure is formalized in Algorithm~\ref{alg:pred}.

\begin{algorithm}[h]
	\caption{Predominant Motion} 
	\label{alg:pred}
	\begin{algorithmic}[1]
		\STATE $\duvec{\pO\pW_F}^w \est \frac{1}{2} \tonde{\duvec{\pO\pW_1}^w + \duvec{\pO\pW_2}^w}$
		\STATE $\duvec{\pO\pW_R}^w \est \frac{1}{2} \tonde{\duvec{\pO\pW_3}^w + \duvec{\pO\pW_4}^w}$
		\IF{$\duvec{\pO\pW_F}^w \cdot \duvec{\pO\pW_R}^w \le 0$}
			\STATE rotation is predominant
			\IF{$\dot{\yaw}^\star > 0$}
				\STATE counterclockwise motion is predominant
			\ELSE
				\STATE clockwise motion is predominant
			\ENDIF
		\ELSE
			\STATE translation is predominant
			\STATE $\alpha \est \atandue{v_{lw}^\star, v_{fw}^\star}$
			\STATE $\alpha_i \est \atandue{\uj^\top \uvec{\pK\pW_i}^s, \ui^\top \uvec{\pK\pW_i}^s} \forevery i \in \quadre{1:4}$
			\IF{$\alpha \in \quadre{\alpha_2; \alpha_1}$}
				\STATE forward motion is predominant
			\ELSIF{$\alpha \in \quadre{\alpha_1; \alpha_3}$}
				\STATE leftward motion is predominant
			\ELSIF{$\alpha \in \quadre{\alpha_4; \alpha_2}$}
				\STATE rightward motion is predominant
			\ELSIF{$\alpha \in \quadre{-\pi; \alpha_4} \cup \quadre{\alpha_3; \pi}$}
				\STATE backward motion is predominant
			\ENDIF
		\ENDIF
	\end{algorithmic}
\end{algorithm}

Since the gait schedule always starts with $\gamma\di{1} = 1$, the gait schedule is updated only when $t = 0$ to avoid inconsistencies that could make the robot fall.

\subsubsection{Trajectories Generator} \label{sssec:trajgen}

This block computes the position $\uvec{\pO\pF}_{1:4}^{w\star}$, velocity $\duvec{\pO\pF}_{1:4}^{w\star}$, and acceleration $\dduvec{\pO\pF}_{1:4}^{w\star}$ references for the feet, using all the data computed by the previous blocks.
Specifically, the gait schedule $\gamma$ is used to compute the lift-off times $t_{0,1:4}$ by Equation~\eqref{eq:liftofftimes}, the period $T$, the current positions $\uvec{\pO\pW}_{1:4}^w$, and the desired velocities $\uvec{\pO\pW}_{1:4}^{w\star}$ of the imaginary wheels (needed to compute the footholds).

As already mentioned in Section~\ref{sssec:soatraj}, after computing the footholds, it is necessary to connect them with a trajectory that lifts each foot $i$ from the ground at time $t_{0,i}$, raises it to the height $h$, and places it back on the ground at time $t_{0,i} + T_{sw}$.
During the rest of the period, the foot must remain stationary.
This means that, as seen in Section~\ref{ssec:kinematics}, since each foot $i$ at time $t_{0,i}$ and at time $t_{0,i} + T_{sw}$ touches the ground, it must hold
\begin{equation} \label{eq:OFreq}
	\begin{cases}
		\duvec{\pO\pF_i}_\star^w\di{t_{0,i}} = \duvec{\pO\pF_i}_\star^w\di{t_{0,i} + T_{sw}} = \uoh\\
		\dduvec{\pO\pF_i}_\star^w\di{t_{0,i}} = \dduvec{\pO\pF_i}_\star^w\di{t_{0,i} + T_{sw}} = \uoh
	\end{cases}.
\end{equation}

A normalized cycloid trajectory is chosen for foot $i$, such that
\begin{multline} \label{eq:cycloid}
	\uvec{\pO\pF_i}_\star^w\di{t} = \uvec{\pO\pW_i}^w\di{0} + {}
	\\ {} + \frac{\eta_i\di{t} - \sin \eta_i\di{t}}{2\pi} \duvec{\pO\pW_i}_\star^w T + \frac{1 - \cos \eta_i\di{t}}{2} h \uk
\end{multline}
\begin{equation} \label{eq:dcycloid}
	\duvec{\pO\pF_i}_\star^w\di{t} = \dot{\eta}_i\di{t} \diff{\uvec{\pO\pF_i}_\star^w}{\eta_i}\di{t}
\end{equation}
\begin{equation} \label{eq:ddcycloid}
	\dduvec{\pO\pF_i}_\star^w\di{t} = \ddot{\eta}_i\di{t} \diff{\uvec{\pO\pF_i}_\star^w}{\eta_i}\di{t} + \dot{\eta}_i^2\di{t} \diff{^2\uvec{\pO\pF_i}_\star^w}{\eta_i^2}\di{t}
\end{equation}
where
\begin{equation}
	\diff{\uvec{\pO\pF_i}_\star^w}{\eta_i}\di{t} = \frac{1 - \cos \eta_i\di{t}}{2\pi} \duvec{\pO\pW_i}_\star^w T + \frac{\sin \eta_i\di{t}}{2} h \uk
\end{equation}
and
\begin{equation}
	\diff{^2\uvec{\pO\pF_i}_\star^w}{\eta_i^2}\di{t} = \frac{\sin \eta_i\di{t}}{2\pi} \duvec{\pO\pW_i}_\star^w T + \frac{\cos \eta_i\di{t}}{2} h \uk.
\end{equation}
For Equations~\eqref{eq:cycloid}, \eqref{eq:dcycloid}, and \eqref{eq:ddcycloid} to satisfy the constraints of Equation~\eqref{eq:OFreq}, it can be proved that it must hold
\begin{equation} \label{eq:etareq}
	\begin{cases}
		\eta_i\di{t_{0,i}} = 0\\
		\eta_i\di{t_{0,i} + T_{sw}} = 2\pi\\
		\dot{\eta}_i\di{t_{0,i}} = \dot{\eta}_i\di{t_{0,i} + T_{sw}} = 0
	\end{cases}.
\end{equation}
It can be proved that the choice of $\eta_i\di{t}$ that satisfies Equation~\eqref{eq:etareq} is
\begin{equation}
	\eta_i\di{t} = 2 \pi \tonde{3 s_i^2\di{t} - 2 s_i^3\di{t}} \tc s_i\di{t} = \frac{t-t_{0,i}}{T_{sw}}.
\end{equation}

\subsubsection{Feet Controller} \label{sssec:feetctrl}

This block computes the feet accelerations $\dduvec{\pO\pF}_{1:4}^w$ such that the feet follow the references computed by the \textit{Trajectories Generator} block.

Defining the error on the outputs of foot $i$ as $\wt{\uvec{\pO\pF_i}}^w \est \uvec{\pO\pF_i}_\star^w - \uvec{\pO\pF_i}^w$, 
the feet are controlled to follow the desired trajectory, by imposing the following control law:
\begin{equation} \label{eq:feetctrl}
	\dduvec{\pO\pF_i}^w = \dduvec{\pO\pF_i}_\star^w + k_{p,i} \wt{\uvec{\pO\pF_i}}^w + k_{d,i} \dwt{\uvec{\pO\pF_i}}^w \forevery i \in \graffe{1:4},
\end{equation}
where $k_{p,i}, k_{d,i} \in \bR$ are tuned as PID gains, and $\uvec{\pO\pF_i}^w$ and $\duvec{\pO\pF_i}^w$ are provided by the \textit{Body Manager}.

\subsection{Body Reference} \label{ssec:bodyref}

This block receives the high-level velocity commands (longitudinal velocity $v_{fw}^\star$, lateral velocity $v_{lw}^\star$, yaw rate $\dot{\yaw}^\star$, height $z^\star$, roll $\roll^\star$, pitch $\pitch^\star$), and converts them into references for the body $\uq_0^\star$ and $\duq_0^\star$, which will be forwarded to the \textit{Body Manager}.

Since the robot oscillates while walking, the velocity references are numerically integrated to obtain a position reference robust to oscillations.
In this way, the robot returns to the desired trajectory even if its position drifts because of the oscillations.
More specifically, $\yaw^\star$ is obtained by integrating $\dot{\yaw}^\star$ and $\uvec{\pO\pG}_\star^w$ is obtained by integrating $\duvec{\pO\pG}_\star^w = R_z\di{\yaw^\star} \bigtonde{v_{fw}^\star \ui + v_{lw}^\star \uj}$.
At this point, $\uvec{\pO\pZ}_\star^w$ can be computed as a function of $\uphi^\star = \matrice{\roll^\star & \pitch^\star & \yaw^\star}^\top$, $\duphi^\star = \matrice{0 & 0 & \dot{\yaw}^\star}^\top$ and $\uvec{\pO\pG}_\star^w$ by applying Equation~\eqref{eq:zmp9}.

\begin{figure}[h]
	\centering%
	\includegraphics[width=0.8\linewidth]{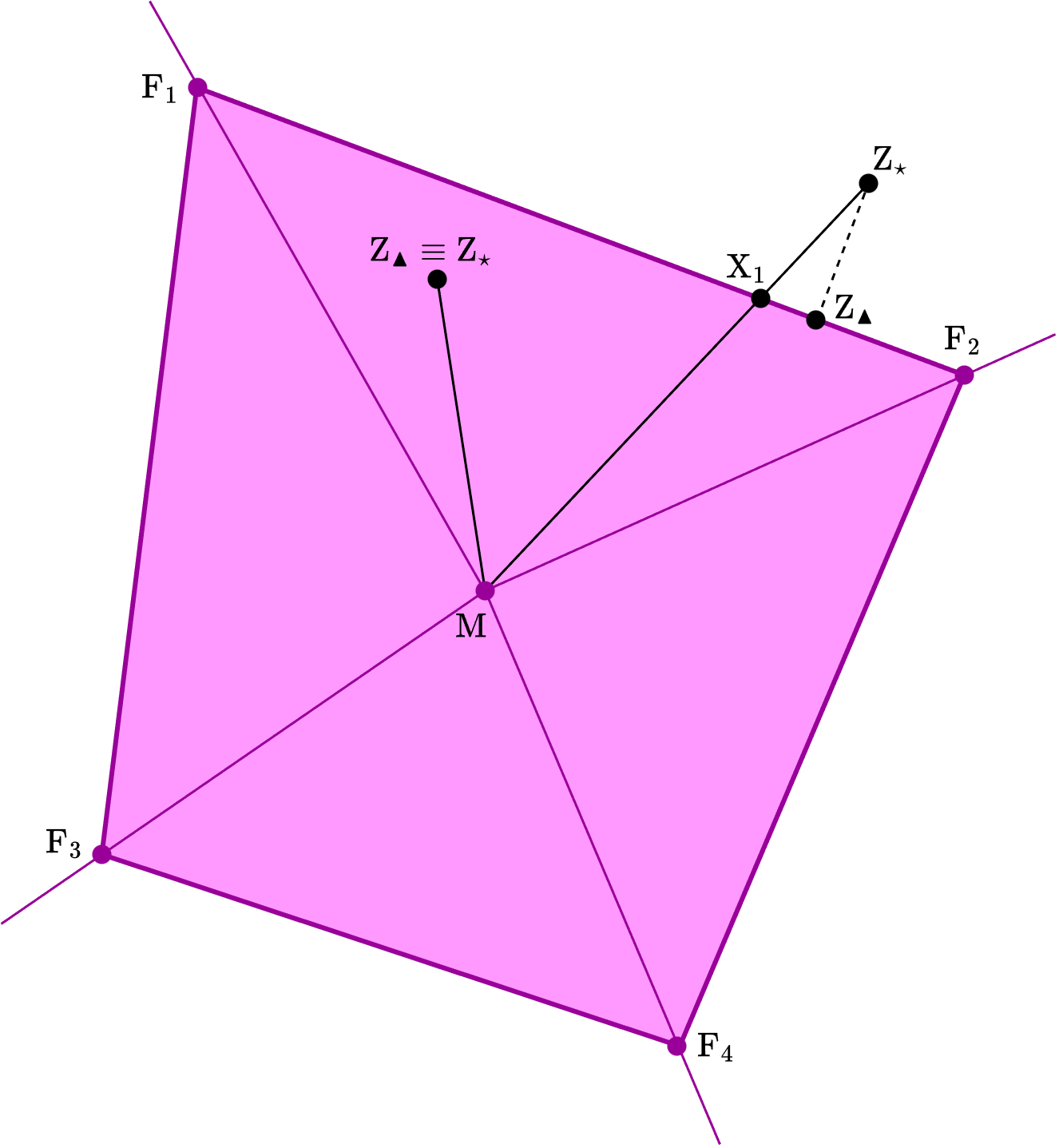}
	\caption{Choice of $\pZ_\blacktriangle$ when $\pZ_\star$ is inside or outside the Support Polygon, with 4 feet on the ground.}
	\label{fig:OZ_r}
\end{figure}

Since it is not certain that the computed $\uvec{\pO\pZ}_\star^w$ belongs to the Support Polygon, we compute
\begin{equation} \label{eq:OZ_safe}
    \uvec{\pO\pZ}_\blacktriangle^w = \arg \min \graffe{\norm{\uvec{\pO\pP}^w - \uvec{\pO\pZ}_\star^w} \tc \uvec{\pO\pP}^w \in \sP_{sup}}
\end{equation}
that is the point belonging to the Support Polygon closest to $\uvec{\pO\pZ}_\star^w$ (see Figure~\ref{fig:OZ_r}).

Inverting Equation~\eqref{eq:zmp9}, we compute
\begin{equation}
    \uvec{\pO\pG}_\blacktriangle^w = \uvec{\pO\pZ}_\blacktriangle^w + \frac{z^\star \uF_{Vg}^{w\star} - \uk \times \uM_V^{w\star}}{\uk^\top \uF_{Vg}^{w\star}}
\end{equation}
which is the position of the center of mass closest to the desired $\uvec{\pO\pG}_\star^w$ that ensures that the ZMP belongs to the Support Polygon, i.e., that the robot is stable. Finally, $\uq_0^\star = \matrice{\uvec{\pO\pG}_\blacktriangle^w \\ \uphi^\star}$ and $\duq_0^\star = \matrice{\duvec{\pO\pG}_\star^w \\ \duphi^\star}$ are forwarded to the \textit{Body Manager}.

\subsection{Body Manager} \label{ssec:bodyman}

\begin{figure}[h]
	\centering%
	\includegraphics[width=\linewidth]{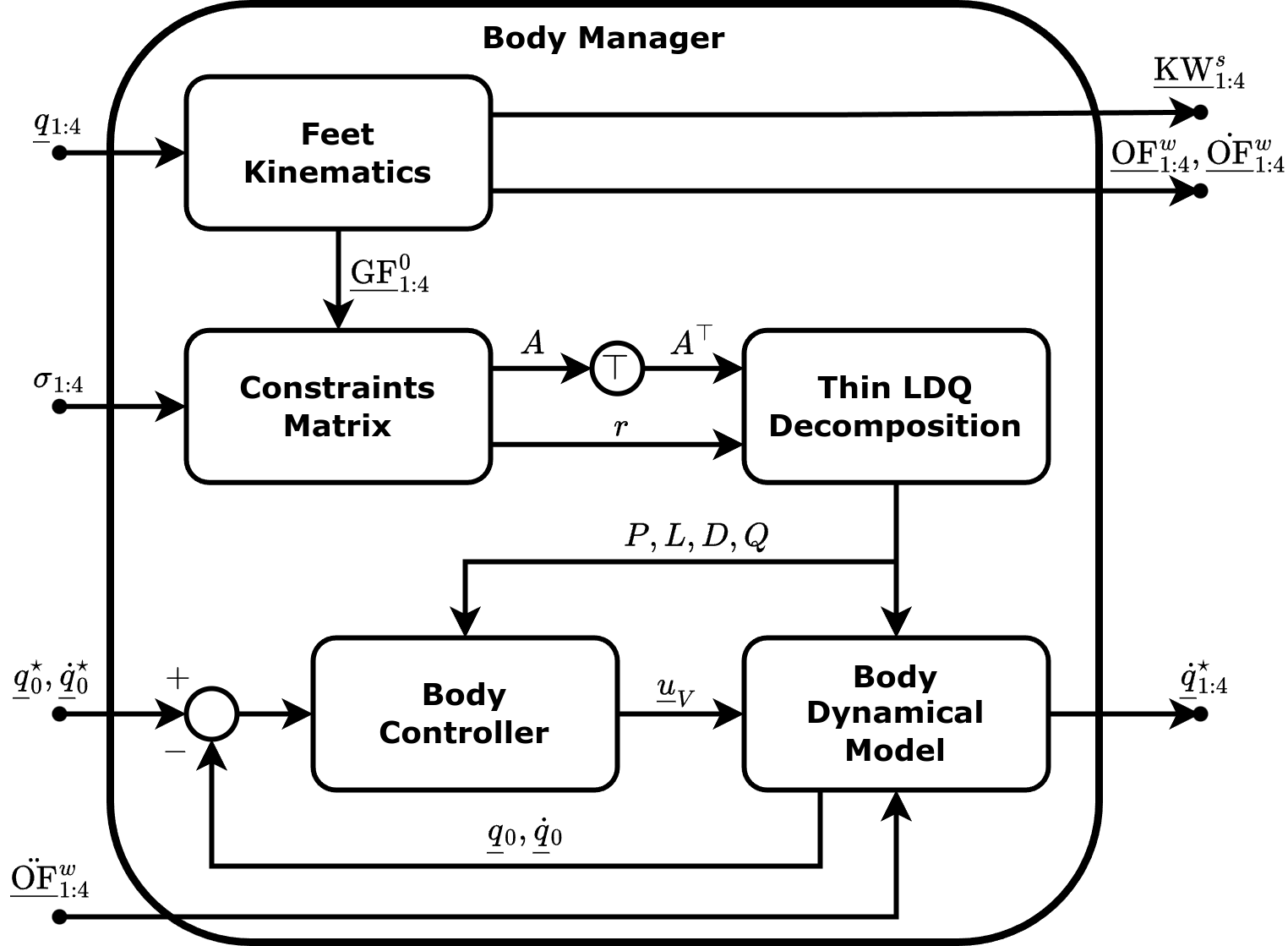}%
	\caption{Body manager block diagram.}%
	\label{fig:bodybd}
\end{figure}

This block receives the feet accelerations $\dduvec{\pO\pF}_{1:4}^w$ from the \textit{Feet Manager}, the body references from the \textit{Body Reference}, the joint positions $\uq_{1:4}$ and the contact signals $\sigma_{1:4}$ from the robot's sensors, and derives the joint velocity commands $\duq_{1:4}^\star$ that will be forwarded to the robot's actuators.
As illustrated in Figure \ref{fig:bodybd}, the Body Manager comprises five blocks described below.

\subsubsection{Feet Kinematics}

This block computes:
the positions $\uvec{\pO\pF}_{1:4}^w$ and velocities $\duvec{\pO\pF}_{1:4}^w$ of the foot contact points in $\SdR{W}$, needed by the \textit{Feet Controller} (see Section~\ref{sssec:feetctrl}), applying Equations~\eqref{eq:OF} and \eqref{eq:dOF};
the positions $\uvec{\pK\pW}_{1:4}^s$ of the imaginary wheels in $\SdR{S}$, needed by the \textit{Wheels Commands} block (see Section~\ref{sssec:wheelcmds});
and the positions $\uvec{\pG\pF}_{1:4}^0$ of the foot contact points in $\SdR{B_0}$, needed by the \textit{Constraints Matrix} block (see Section~\ref{sssec:constmatrix}), applying Equation~\eqref{eq:GF}.

\subsubsection{Constraints Matrix} \label{sssec:constmatrix}

This block uses the contact signals $\sigma_{1:4}$ and the positions of the contact points $\uvec{\pG\pF}_{1:4}^0$ in the frame $\SdR{B_0}$ to compute the matrix $A$, applying Equation~\eqref{eq:matrixA}, and its rank $r$, according to Table~\ref{tab:r}.
This data is then forwarded to the \textit{Thin LDQ Decomposition} block (see Section~\ref{sssec:ldq}).

\subsubsection{Thin LDQ Decomposition} \label{sssec:ldq}

This block decomposes matrix $A^\top$ into the matrices $P$, $L$, $D$, $Q$ such that $A^\top = PLDQ$.
These data are necessary for the \textit{Body Dynamical Model} (see Section~\ref{sssec:bodymodel}) and the \textit{Body Controller} (see Section~\ref{sssec:bodyctrl}) to work.

As proved by Golub and Van Loan~\cite[Theorem 3.2.1]{golub2013matrix}, for any square matrix $S \in \bR^{n \times n}$ with rank $r$, there exists a unique permutation matrix $P \in \graffe{0,1}^{n \times n}$, a unique unit lower triangular matrix $L \in \bR^{n \times r}$, a unique unit upper triangular matrix $U \in \bR^{r \times n}$, and a unique diagonal matrix $D \in \bR^{r \times r}$ such that $S = P L D U P^\top$.
This is known as \textit{LDU Decomposition}.
Assuming that row and column indices start from 0, the matrices $P$, $L$, $D$, and $U$ are computed as follows:
\begin{equation} \label{eq:ldu}
	\begin{cases}
		\up_k \in \graffe{\ui_0, ..., \ui_{n-1}} \smallsetminus \graffe{\up_0, ..., \up_{k-1}}\\
		\calcbw{\ell_{kj} = \frac{1}{d_j} \tonde{\up_k^\top S \up_j - \sum_{i=0}^{j-1} \ell_{ki} d_i u_{ij}}}{j=0}{k-1} \\
		\calcbw{u_{jk} = \frac{1}{d_j} \tonde{\up_j^\top S \up_k - \sum_{i=0}^{j-1} \ell_{ji} d_i u_{ik}}}{j=0}{k-1} \\
		d_k = \up_k^\top S \up_k - \sum_{i=0}^{k-1} \ell_{ki} d_i u_{ik} 
	\end{cases}
\end{equation}
for each $k$ from 0 to $r-1$, where $\graffe{\ui_0, ..., \ui_{n-1}}$ is the canonical basis of $\bR^n$.
To make the decomposition robust to numerical errors, $\up_k$ is chosen to maximize $\abs{d_k}$.
This is implemented with a nested \textsf{for} loop that always takes $nr - \frac{r\tonde{r-1}}{2}$ iterations to derive $P$, $L$, $U$ and $D$.
This aspect is important when it comes to characterizing the predictability of the system during a real-time analysis.

When $S$ is nonsingular, i.e., when $r = n$, the matrices $P$, $L$, $D$, and $U$ are fully computed by the algorithm in Equation~\eqref{eq:ldu}, and it holds that $S^{-1} = P U^{-1} D^{-1} L^{-1} P^\top$.
This is useful because $U^{-1}$, $L^{-1}$, and $D^{-1}$ are much easier to compute than $S^{-1}$.

When $S$ is singular, i.e., when $r < n$, it results in $P = \matrice{P_{r-1} & P_+}$, $L = \matrice{L_{r-1} \\ L_+}$, and $U = \matrice{U_{r-1} & U_+}$, where $P_{r-1}$, $L_{r-1}$, $D$, and $U_{r-1}$ are computed by the algorithm in Equation~\eqref{eq:ldu}, while the columns of $P_+$ are $\graffe{\ui_0, ..., \ui_{n-1}} \smallsetminus \graffe{\up_0, ..., \up_{r-1}}$, $L_+ = P_+^\top S P_{r-1} U_{r-1}^{-1} D^{-1}$, and $U_+ = D^{-1} L_{r-1}^{-1} P_{r-1}^\top S P_+$.

As proved by Golub and Van Loan~\cite[Theorem 4.1.3]{golub2013matrix}, as a consequence of the previous result, if $S$ is symmetric, that is, $S = S^\top$, then $U = L^\top$ and the LDU decomposition can be simplified to $S = P L D L^\top P^\top$.
This is known as \textit{LDLT Decomposition}.

Let $X \in \bR^{m \times n}$ be a matrix with rank $r \le \min\graffe{m,n}$.
For the \textit{LDLT Decomposition}, since matrix $S = X X^\top$ is symmetric and positive definite by construction, then $X X^\top = P L D_S L^\top P^\top$, where $P \in \graffe{0,1}^{m \times m}$ is a permutation matrix, $L \in \bR^{m \times r}$ is a unit lower triangular matrix, and $D_S \in \bR^{r \times r}$ is a diagonal matrix.

As proved by Golub and Van Loan~\cite[Theorem 5.2.3]{golub2013matrix},
there exists a matrix $Q \in \bR^{r \times n}$ such that $Q Q^\top = I_r$ and a matrix $D$ such that $D^2 = D_S$, for which $X = P L D Q$, in such a way that $X X^\top = P L D Q Q^\top D L^\top P^\top = P L D_S L^\top P^\top$.
It can be proved that $Q = D^{-1} L_{r-1}^{-1} P_{r-1}^\top X$.

If we apply this results with $X = A^\top$, we obtain the matrices $P \in \graffe{0,1}^{6 \times 6}$, $L \in \bR^{6 \times r}$, $D \in \bR^{r \times r}$, and $Q \in \bR^{r \times 12}$ such that
\begin{equation} \label{eq:PLDQ}
	A^\top = P L D Q.
\end{equation}
This result is crucial, as it will be employed to find a solution to Equation~\eqref{eq:invcdyn}.

\subsubsection{Body Dynamical Model} \label{sssec:bodymodel}

This block internally simulates the dynamic system to be controlled, using the matrix decomposition of Equation~\eqref{eq:PLDQ} and integrating the feet acceleration commands $\dduvec{\pO\pF}_{1:4}^w$ incoming from the \textit{Feet Manager} block.
The internal simulation allows the computation of the data necessary for the \textit{Body Controller} (see Section~\ref{sssec:bodyctrl}) and \textit{Body Reference} (see Section~\ref{ssec:bodyref}) blocks to operate, and the joint velocity commands $\duq_{1:4}^\star$ that make the robot move.
The internal simulation is based on the internal model explained in the following paragraphs.

From Equation~\eqref{eq:dyn} we derive
\begin{equation}
	M_\pG^0 \duV_{w0}^0 - \sum_{j=1}^4 \sigma_i A_i^\top \uF_i^0 = \uW_g^0 - \quadre{\ads \uV_{w0}^0} M_\pG^0 \uV_{w0}^0
\end{equation}
From Equation~\eqref{eq:grf2} we derive
\begin{equation}
	- \sigma_i A_i \duV_{w0}^0 = \sigma_i \tonde{\uh_i + \dduvec{\pG\pF_i}^0}
\end{equation}
Combining such results together we obtain
\begin{equation} \label{eq:invcdyn}
	\underbrace{\matrice{M_\pG^0 & -A^\top \\ -A & O}}_{\Delta} \matrice{\duV_{w0}^0 \\ \uF_{1:4}^0} = \matrice{ \uh_V \\ \Sigma \tonde{\uh_{1:4} + \dduvec{\pG\pF}_{1:4}^0}}
\end{equation}
where we have defined 
$\uh_V \est \uW_g^0 - \quadre{\ads \uV_{w0}^0} M_\pG^0 \uV_{w0}^0$.

We observe that $\Delta \in \bR^{18 \times 18}$ is not invertible because $\rank A = r \le 6 \THEN \rank \Delta = r+6 \le 12 < 18$. 
However, after substituting Equation~\eqref{eq:PLDQ} into Equation~\eqref{eq:invcdyn}, we derive the first derivative of the twist
\begin{equation} \label{eq:duV}
	\duV_{w0}^0 = \Lambda \uh_V + \Psi D^{-1} Q \tonde{\uh_{1:4} + \dduvec{\pG\pF}_{1:4}^0}
\end{equation}
where
\begin{equation} \label{eq:PsiFF}
	\Phi = - \quadre{L^\top P^\top \quadre{M_\pG^0}^{-1} PL}^{-1}
\end{equation}
\begin{equation} \label{eq:PsiVF}
	\Psi = \quadre{M_\pG^0}^{-1} PL \Phi
\end{equation}
\begin{equation} \label{eq:PsiVV}
	\Lambda = \quadre{M_\pG^0}^{-1} \quadre{I_6 + PL \Psi^\top}
\end{equation}

Let's combine the previously seen equations to derive a state form to be controlled: from Equation~\eqref{eq:twist} we get
\begin{equation} \label{eq:duq0}
	\duq_0 = \quadre{J_{w0}^0\di{\uphi}}^{-1} \uV_{w0}^0
\end{equation}
From Equation~\eqref{eq:dGF} we get
\begin{equation} \label{eq:duqj}
	\duq_i = \quadre{J_{\pG\pE_i}^0\di{\uq_i}}^{-1} \tonde{\duvec{\pG\pF_i}^0 - \uom_{w0}^0 \times \zeta \quadre{R_0^w}^\top \uk}.
\end{equation}
From Equation~\eqref{eq:ddOF} we get
\begin{equation} \label{eq:ddGF}
	\dduvec{\pG\pF_i}^0 = - \uh_i - A_i \duV_{w0}^0 + \quadre{R_0^w}^\top \dduvec{\pO\pF_i}^w.
\end{equation}
Since the robot's body has only $r$ controllable degrees of freedom, as explained in Section~\ref{sssec:uncm}, the control inputs for the body are chosen as $\uu_V \est Q \dduvec{\pG\pF}_{1:4}^0 \in \bR^r$,
which are the projections of the foot accelerations in $\SdR{B_0}$ into the controllable motions space.

Hence, from Equation~\eqref{eq:duV} we get 
\begin{equation} \label{eq:duV2}
	\duV_{w0}^0 = \uf_V + \Psi D^{-1} \uu_V
\end{equation}
where $\uf_V \est \Lambda \uh_V + \Psi D^{-1} Q \uh_{1:4}$ is defined.
Consequently, combining Equations~\eqref{eq:duq0}, \eqref{eq:duqj}, \eqref{eq:ddGF}, and \eqref{eq:duV2}, we obtain the following state form
\begin{equation} \label{eq:stateform}
	\begin{cases}
		\duq_0 &= \quadre{J_{w0}^0\di{\uphi}}^{-1} \uV_{w0}^0 \\
		\duq_i &= \quadre{J_{\pG\pE_i}^0\di{\uq_i}}^{-1} \tonde{\duvec{\pG\pF_i}^0 - \uom_{w0}^0 \times \zeta \quadre{R_0^w}^\top \uk} \\
		\duV_{w0}^0 &= \uf_V + \Psi D^{-1} \uu_V \\
		\dduvec{\pG\pF_i}^0 &= - \uh_i - A_i \tonde{\uf_V + \Psi D^{-1} \uu_V} + \quadre{R_0^w}^\top \dduvec{\pO\pF_i}^w 
	\end{cases}
\end{equation}
for every $i \in \graffe{1:4}$.
Thus the vectors $\uq_0 \in \bR^6$, $\uq_{1:4} \in \bR^{12}$, $\uV_{w0}^0 \in \bR^6$, and $\duvec{\pG\pF}_{1:4}^0 \in \bR^{12}$ constitute the state $\ux \in \bR^{36}$ of the system, while the vectors $\uu_V \in \bR^r$ and $\dduvec{\pO\pF}_{1:4}^w \in \bR^{12}$ constitute the inputs $\uu \in \bR^{r+12}$ of the system.

\subsubsection{Body Controller} \label{sssec:bodyctrl}

This block computes the control inputs $\uu_V$ (see Equation~\eqref{eq:duV2}) that make the body state $\uq_0$ and $\duq_0$ (see Section~\ref{sssec:bodymodel}) follow the body references $\uq_0^\star$ and $\duq_0^\star$ (see Section~\ref{ssec:bodyref}), i.e., the inputs that bring the errors $\wt{\uq}_0 \est \uq_0^\star - \uq_0$ and $\dwt{\uq}_0 \est \duq_0^\star - \duq_0$ to zero.

By differentiating Equation~\eqref{eq:twist}, we obtain
\begin{equation} \label{eq:dtwist}
	\duV_{w0}^0 = \dot{J}_{w0}^0\di{\uphi,\duphi} \duq_0 + J_{w0}^0\di{\uphi} \dduq_0,
\end{equation}
from which, combining it with Equation~\eqref{eq:duV2}, we derive
\begin{equation}
	\dduq_0 = \quadre{J_{w0}^0\di{\uphi}}^{-1} \tonde{\uf_V + \Psi D^{-1} \uu_V - \dot{J}_{w0}^0\di{\uphi,\duphi} \duq_0}
\end{equation}
Now, applying the \textit{Feedback Linearization} technique, we find a single control law valid for any value of $r$:
\begin{equation} \label{eq:crtlf}
	\uu_V = - D L^\top P^\top \bigtonde{J_{w0}^0 \bigtonde{k_{p,0} \wt{\uq}_0 + k_{d,0} \dwt{\uq}_0} + \dot{J}_{w0}^0 \duq_0 - \uf_V}
\end{equation}
where $k_{p,0}, k_{d,0} \in \bR$ are tuned as PID gains.

The advantages resulting from this control law make this formulation one of the key contributions of this work:
(i) it is a closed-form solution, 
(ii) it is valid for any combination of feet contacts,
(iii) it linearizes exactly a complex system like a quadruped robot, and
(iv) it projects the linear control law for the pose into the controllable degrees of freedom granted by the foot contacts.
	\section{Experimental results} \label{sec:results}

This section describes the results of some experiments carried out to test the effectiveness of the proposed approach.

\subsection{Implementation} \label{ssec:impl}

The framework is implemented in C++, following the architecture presented in Section~\ref{sec:framework}.
It does not rely on any third-party libraries to keep the implementation of \NAME\ as predictable and maintainable as possible.

To make this possible, a custom matrix calculation library called \textsc{Laerte} (\textit{Linear Algebra for Embedded Real-Time systEms}) has been implemented.
\textsc{Laerte} uses only static allocation and avoids dynamic memory allocation, which is one of the main causes of temporal unpredictability in real-time systems.
It optimizes the performance of \NAME\ by implementing only the operations necessary for the framework to run, such as matrix addition, multiplication, inversion, and the LDU and LDQ decompositions explained in Section~\ref{sssec:ldq}.

The proposed solution has been tested in simulation, but to obtain more significant results, the positions of the foot center points $\uvec{\pG\pE_i}^0$ (see Equation~\eqref{eq:GE}), the radius of the spherical foot of Equation~\eqref{eq:OF} and the dynamic parameters of Equation~\eqref{eq:dyn} are derived from an actual robot.
The reference platform is a Unitree Go2\footnote{\textsf{https://www.unitree.com/go2}} whose parameters are derived from the Unified Robot Description Format (URDF) file made available by the manufacturer\footnote{\textsf{https://github.com\-/unitreerobotics\-/unitree\_ros\-/blob\-/master\-/robots\-/go2\_description\-/urdf\-/go2\_description.urdf}}.

The following values were chosen for the other parameters:
for the feet workspaces, $\abs{x_{i,min}^s} \est 0.05$ m, $\abs{y_{i,min}^s} \est 0.05$ m, $\abs{x_{i,max}^s} \est 0.5$ m, and $\abs{y_{i,max}^s} \est 0.5$ m (see Equation~\eqref{eq:ws});
for the gait parameters, $T_{sw}^{min} \est 0.2$ s, $T_{sw}^{max} \est 0.2$ s, $\beta_{min} \est 0.5$, and $\beta_{max} \est 0.8$ (see Section~\ref{sssec:period});
for the step height, $h \est 0.05$ m (see Section~\ref{sssec:trajgen});
for the feet controllers, $k_{p,i} \est 1000$ and $k_{d,i} \est 110 \forevery i \in \graffe{1:4}$ (see Equation~\eqref{eq:feetctrl});
for the body controller, $k_{p,0} \est 100$ and $k_{d,0} \est 21$ (see Equation~\eqref{eq:crtlf}).

To validate the control framework, \NAME\ is tested on a quadruped robot modeled by the kinematic and dynamic equations described in Section~\ref{sec:model}, simulating the signals from the robot:
joint angles $\uq_{1:4}$ are obtained by numerically integrating the joint velocity commands $\duq_{1:4}^\star$;
the foot contact signals are obtained by assuming that each foot $i$ is in contact with the ground when the foot contact point $\pF_i$ is below a certain height, that is, $\sigma_i \est \graffe{\uk^\top \uvec{\pO\pF_i}^w \le 5 \cdot 10^{-3}}$.

\begin{figure*}[t]
	\centering%
	\begin{subfigure}{0.7\linewidth}
		\includegraphics[width=\linewidth]{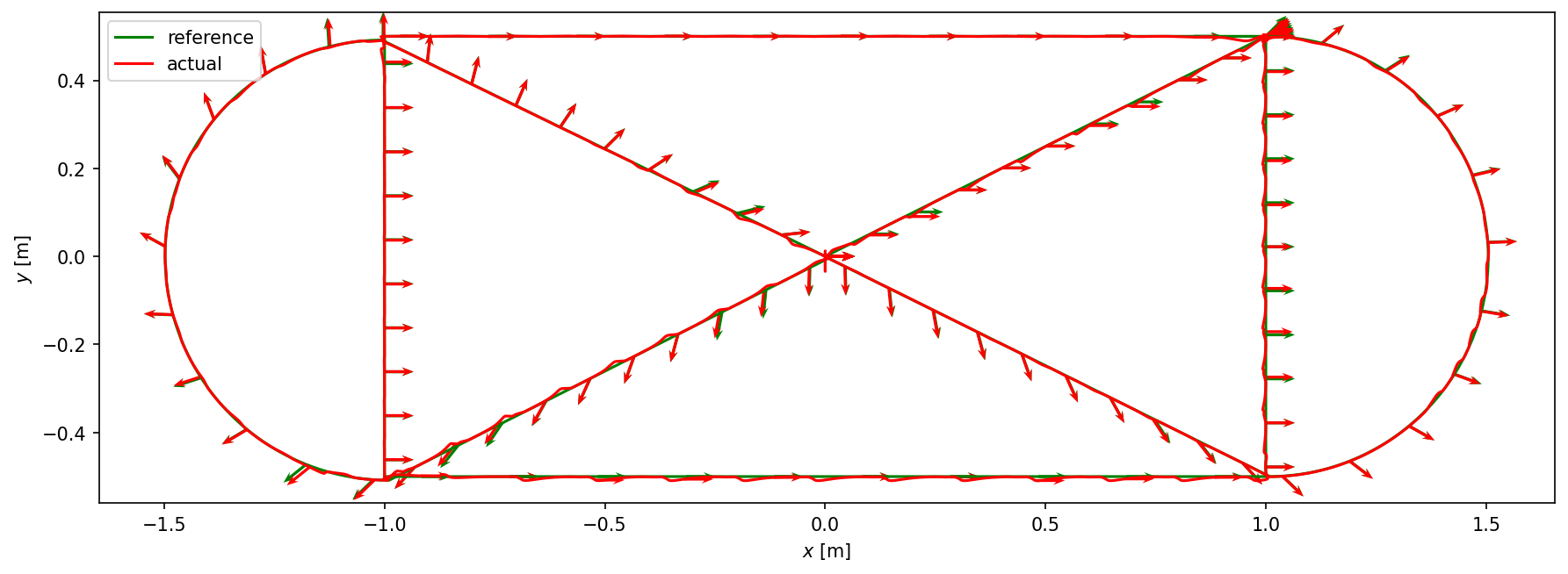}%
		\caption{Reference and actual planar paths for low speeds.}%
		\label{fig:2Dtest:slow}
	\end{subfigure}%
	\\\begin{subfigure}{0.7\linewidth}
		\includegraphics[width=\linewidth]{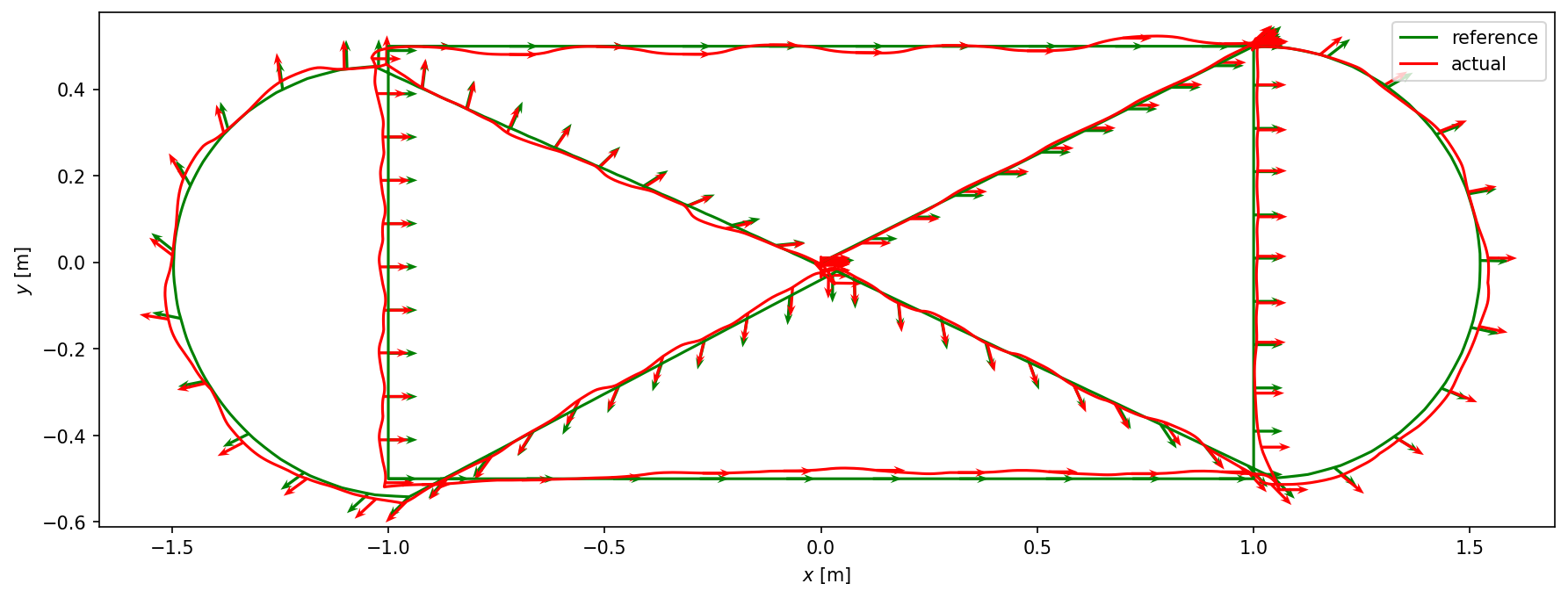}%
		\caption{Reference and actual planar paths for high speeds.}%
		\label{fig:2Dtest:fast}
	\end{subfigure}%
	\caption{Trajectories used for the two experiments executed at low speed (a) and high speed (b), respectively. The green curve indicates the reference path, while
	the red curve indicates the path actually traversed by the robot.
	The arrows indicate the projections of the unit vector of axis $x_0$ of frame $\SdR{B_0}$ onto the plane, to show $\yaw$ values.}
	\label{fig:2Dtest}
\end{figure*}

\subsection{Controller performance test} \label{ssec:expctrl}

To test \NAME's ability to follow a general reference, the robot was commanded to follow the reference trajectory illustrated in Figure~\ref{fig:2Dtest}, which was designed to cover all possible command combinations.

Initially, starting from the center, the robot must raise and lower itself on the spot (control of $z$), also raising and lowering the nose (control of $\pitch$) and the sides (control of $\roll$);
then, it must move to the front-left vertex of the rectangle and traverse the entire perimeter without rotating on itself (control of $v_{fw}$ and $v_{lw}$);
then, it must rotate on the spot (control of $\dot{\yaw}$); 
then, traverse a semicircular section while rotating on itself without lowering or inclining (control of $v_{fw}$, $v_{lw}$ and $\dot{\yaw}$);
after that, it must traverse two diagonal sections while raising, lowering, and rotating on itself (control of $v_{fw}$, $v_{lw}$, $\dot{\yaw}$ and $z$), a semicircular section in which it lowers while inclining nose and sides, and finally it must return to the starting position and orientation (control of $v_{fw}$, $v_{lw}$, $\dot{\yaw}$, $z$, $\roll$ and $\pitch$).

To test the behavior of the controller at different speeds, the trajectory is traversed slowly (0.2 m/s) in a first experiment and quickly (1 m/s) in a second experiment.
The reference trajectory described above is translated into a sequence of commands, that is, $z^\star$, $\roll^\star$, $\pitch^\star$, $v_{fw}^\star$, $v_{lw}^\star$, and $\dot{\yaw}^\star$, that are sent to \NAME\ by an external test module that mimics the behavior of a user or a high-level planner.

\NAME\ was integrated into a ROS 2 node, to prove its compatibility with the most common robotic frameworks, and also to visualize the data computed internally during the experiments.
The node includes two ROS timers: one at a frequency of 100 Hz that runs the framework, and one at a frequency of 20 Hz that publishes the data necessary to animate the robot.
The published data are then displayed in RViz, the 3D visualization tool for ROS 2.

To easily monitor the progress on RViz, these experiments were conducted on a desktop computer with an Intel® Core™ i9-9900 CPU @ 3.10GHz (16 cores), 32 GB of RAM, and an NVIDIA GeForce RTX 2080 SUPER GPU, running Ubuntu 22.04 LTS and ROS 2 Humble.

\begin{figure}[h]
	\includegraphics[width=\linewidth]{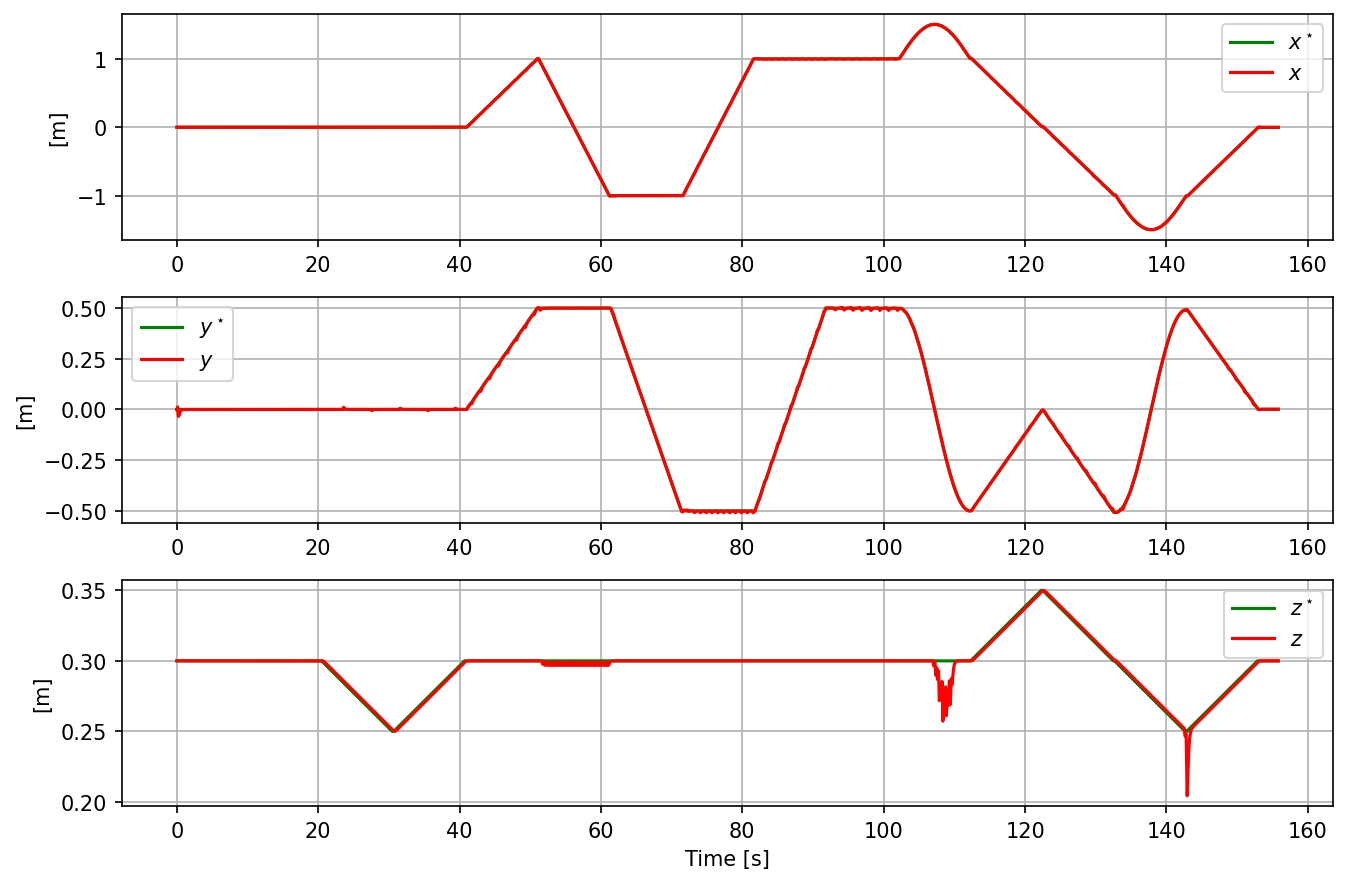}%
	\caption{Time plots of the actual coordinates of the center of mass (red) tracking their corresponding references (green), during the first experiment where the trajectory is traversed slowly.}
	\label{fig:slow:xyz}
\end{figure}

\begin{figure}[h]
	\includegraphics[width=\linewidth]{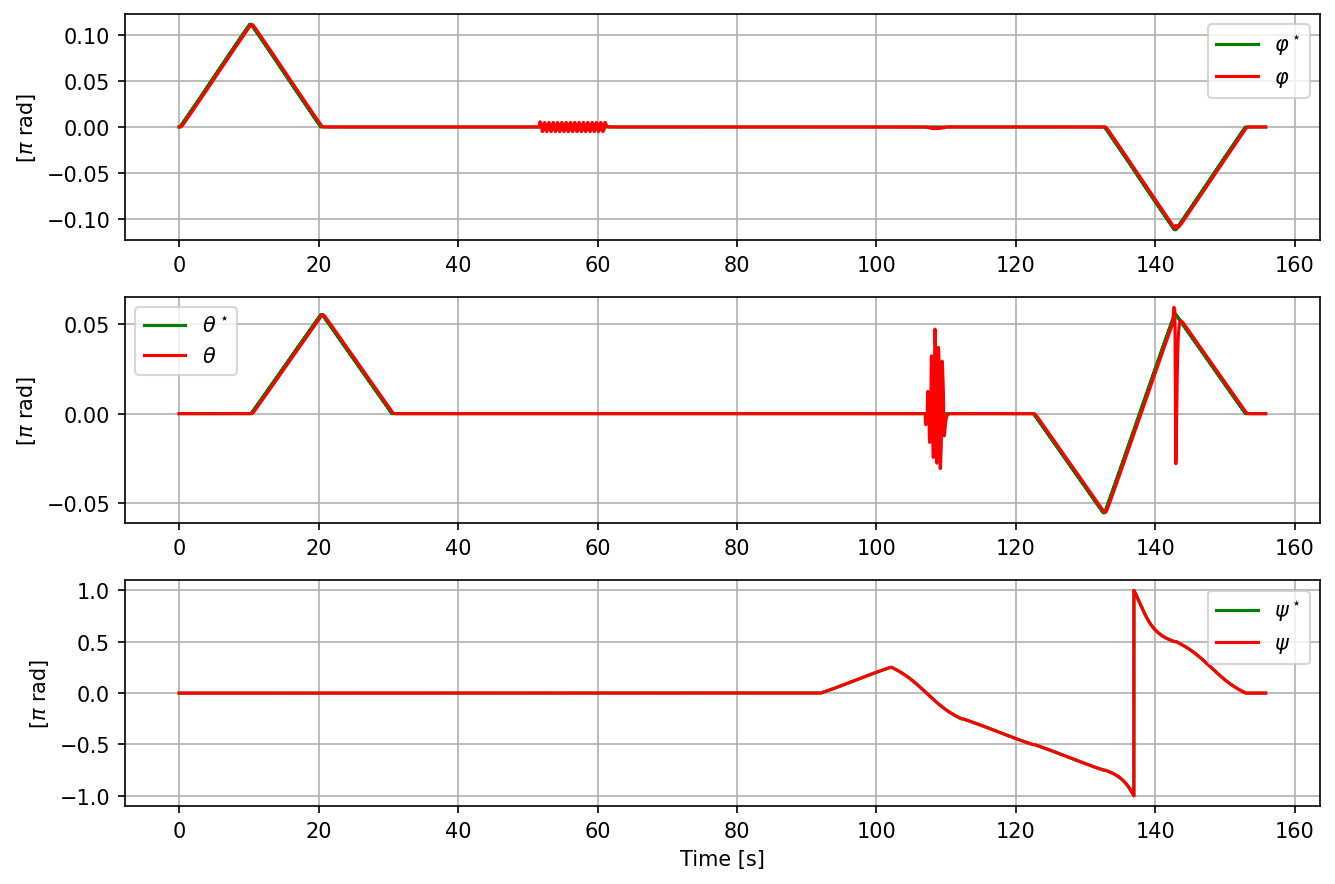}%
	\caption{Time plots of the actual Euler angles of the body (red) tracking their corresponding references (green), during the first experiment where the trajectory is traversed slowly.}
	\label{fig:slow:rpy}
\end{figure}

\begin{figure}[h]
	\includegraphics[width=\linewidth]{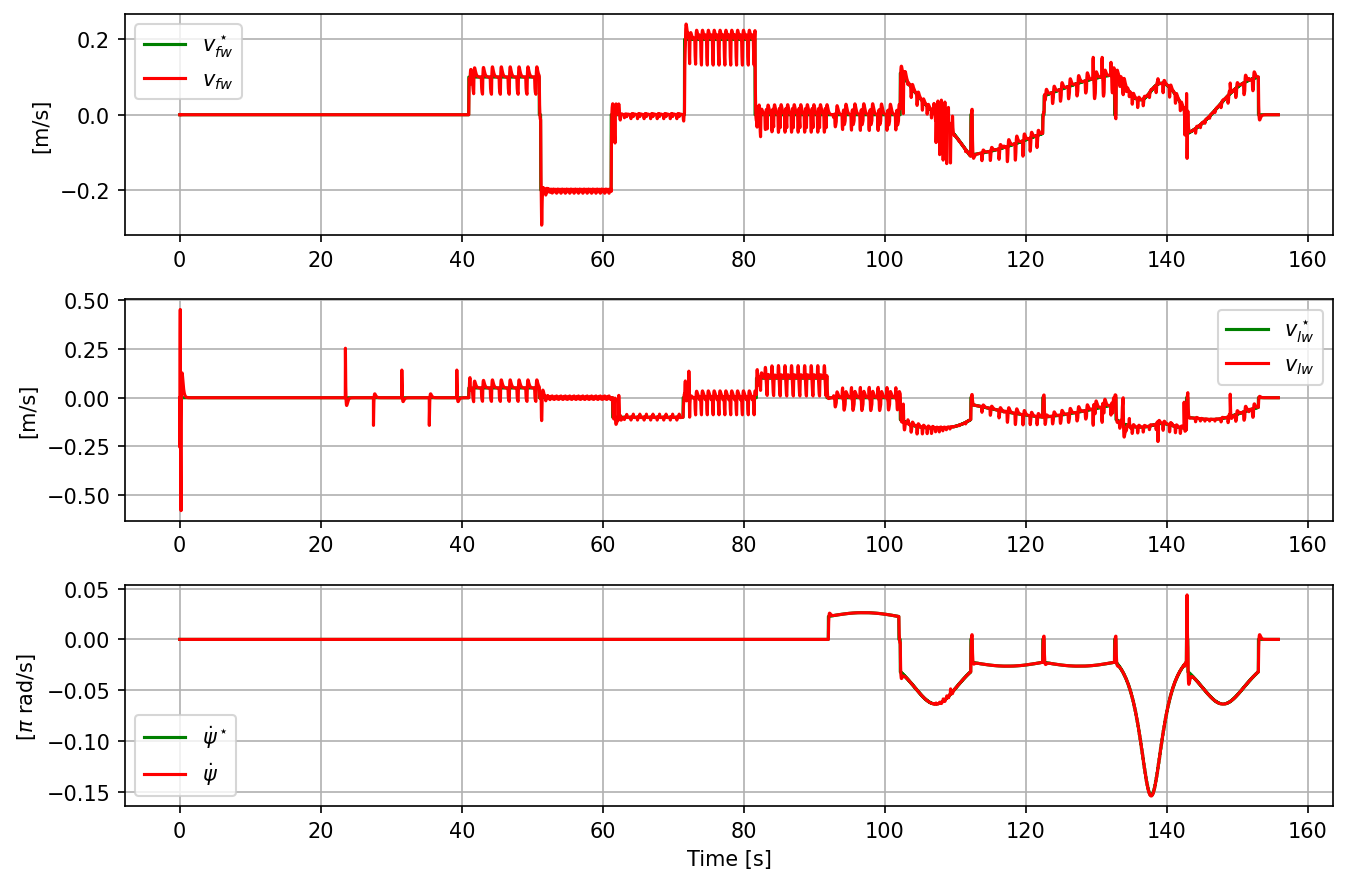}%
	\caption{Time plots of the actual velocities of the body (red) tracking their corresponding references (green), during the first experiment where the trajectory is traversed slowly.}
	\label{fig:slow:vel}
\end{figure}

Figures~\ref{fig:slow:xyz}, \ref{fig:slow:rpy}, and \ref{fig:slow:vel} show the results of the first experiment in which the trajectory is traversed slowly, i.e., with a maximum linear speed of 0.2 m/s and a maximum angular speed of 0.15$\pi$ rad/s.
The plots in Figures~\ref{fig:slow:xyz} and \ref{fig:slow:rpy} show how the values of $x$, $y$, $z$, $\roll$, $\pitch$, and $\yaw$ closely follow their references, except for some oscillation on $z$, $\roll$ and $\pitch$, which is probably due to the gait schedule chosen for that particular combination of commands.
From the experimental data, it can be observed that:
the planar position errors $\wt{x}$ and $\wt{y}$ are below 2 cm;
the height error $\wt{z}$ is generally below 1 cm, with two peaks of around 4 cm;
the orientation errors $\wt{\roll}$, $\wt{\pitch}$ and $\wt{\yaw}$ are generally below 0.005$\pi$ rad, apart from some peaks of less than 0.1$\pi$ rad on $\wt{\pitch}$.
The plots in Figure~\ref{fig:slow:vel} show how the values of $v_{fw}$, $v_{lw}$, and $\dot{\yaw}$ oscillate around the references $v_{fw}^\star$, $v_{lw}^\star$, and $\dot{\yaw}^\star$.
The oscillations around the reference values in this case are due to the fact that the robot, in lifting and landing its feet, creates discontinuities in the model (see Section~\ref{sssec:bodymodel}) and periodically shifts the Support Polygon and then the position of the safe reference for the center of mass $\uvec{\pO\pG}_\blacktriangle^w$ (see Section~\ref{ssec:bodyref}). These values influence the computation of $v_{fw}$ and $v_{lw}$.
Furthermore, from the experimental data it can be observed that, since the velocity references are step inputs, every time the reference changes, there is an impulsive peak in the error due to the reference change and the tracking delay, with a magnitude equal to the step height.
In fact, where the references vary continuously, no error impulses are observed.
Therefore, it can be concluded that for low speeds, the performances of \NAME\ are satisfactory in the control of all six degrees of freedom of the quadruped's body.

\begin{figure}[h]
	\includegraphics[width=\linewidth]{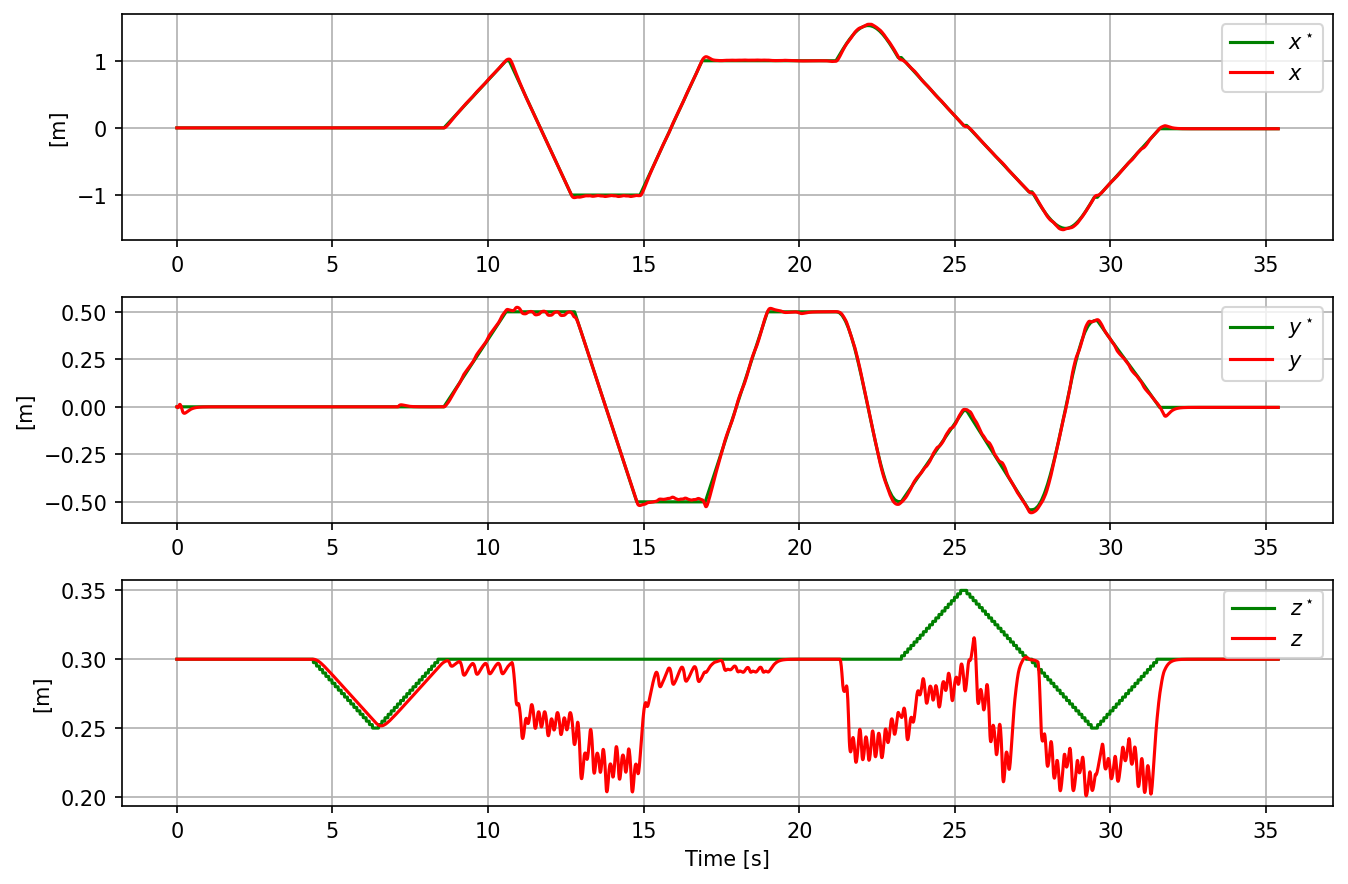}%
	\caption{Time plots of the actual coordinates of the center of mass (red) tracking their corresponding references (green), during the second experiment where the trajectory is traversed quickly.}
	\label{fig:fast:xyz}
\end{figure}

\begin{figure}[h]
	\includegraphics[width=\linewidth]{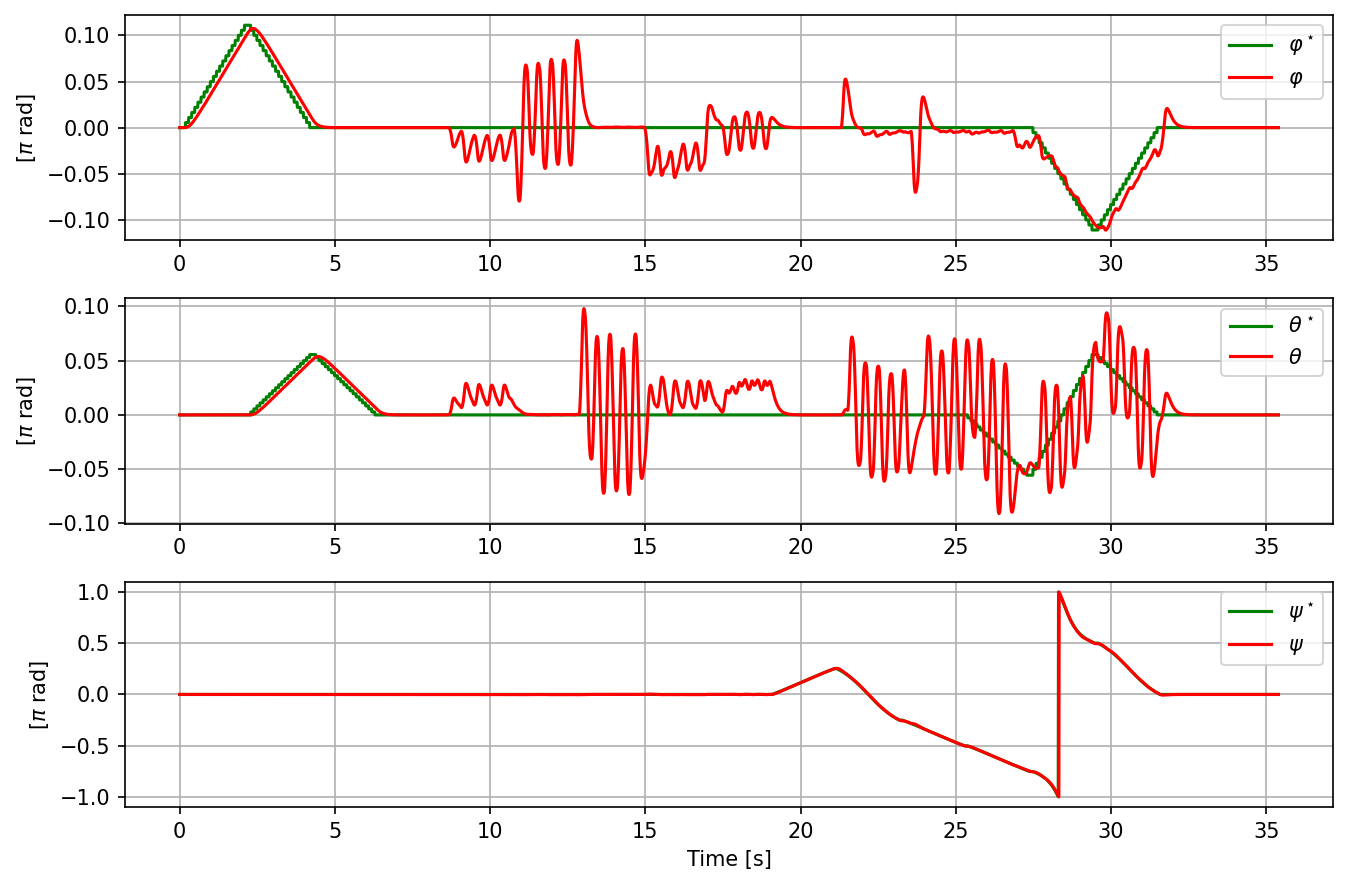}%
	\caption{Time plots of the actual Euler angles of the body (red) tracking their corresponding references (green), during the second experiment where the trajectory is traversed quickly.}
	\label{fig:fast:rpy}
\end{figure}

\begin{figure}[h]
	\includegraphics[width=\linewidth]{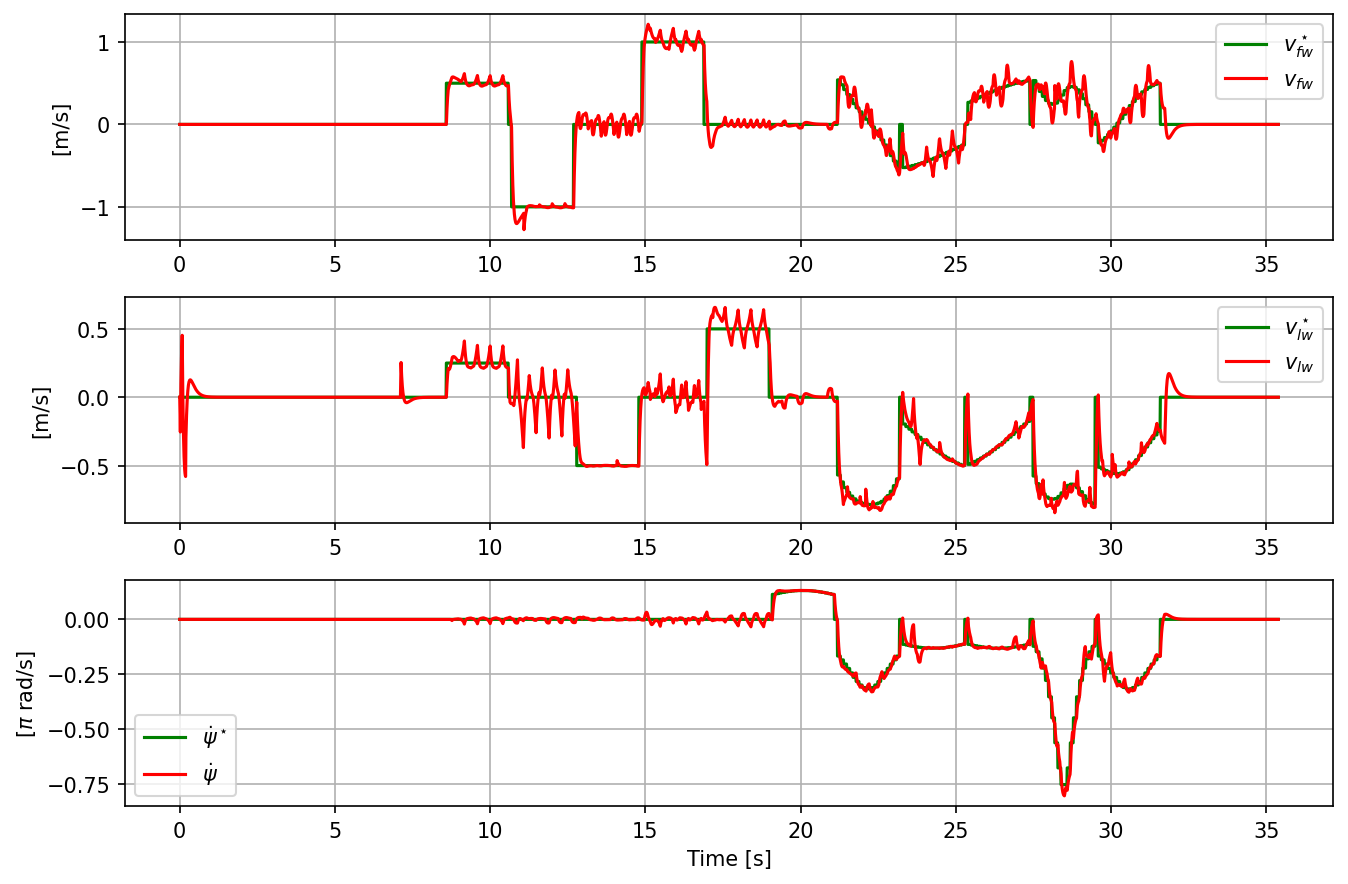}%
	\caption{Time plots of the actual velocities of the body (red) tracking their corresponding references (green), during the second experiment where the trajectory is traversed quickly.}
	\label{fig:fast:vel}
\end{figure}

Figures~\ref{fig:fast:xyz}, \ref{fig:fast:rpy}, and \ref{fig:fast:vel} show the results of the second experiment in which the trajectory is traversed more quickly, that is, with linear speeds up to 1 m/s and angular speeds up to 0.75$\pi$ rad/s.
The plots in Figures~\ref{fig:fast:xyz} and \ref{fig:fast:rpy} show how the values of $x$, $y$ and $\yaw$ still follow the references $x^\star$, $y^\star$ and $\yaw^\star$ quite closely, while the values of $\roll$ and $\pitch$ oscillate considerably around their references $\roll^\star$ and $\pitch^\star$. Also, the value of $z$ struggles in following its reference $z^\star$.
This is due to the fact that, to increase speed, the \textit{Period} block decreases the duty factor $\beta$, as explained in Section~\ref{sssec:period}.
When $\beta \le 0.75$, there are moments in which the robot has less than 3 grounded feet and becomes unstable due to gravity, as explained in Section~\ref{sssec:uncm}.
From the experimental data, it can be observed that: the planar position errors $\wt{x}$ and $\wt{y}$ are below 5 cm, while the height error $\wt{z}$ can reach peaks of 10 cm, and the inclination errors $\wt{\roll}$ and $\wt{\pitch}$ can reach peaks of 0.1$\pi$ rad.
Regarding the velocities, from Figure~\ref{fig:fast:vel} it can be observed that the situation is not very different from the one of the first experiment, for the same reasons.

Overall, this second experiment demonstrates that, when the robot is running, \NAME\ manages to make the robot always follow the high-level velocity commands,
sacrificing the performance of the other commands, that is, the height and inclination of the body.
This is possible because the LDQ decomposition naturally filters out uncontrollable motions, which, due to gravity, coincide with those involving height and inclination of the body.
Hence, it can be concluded that, for high speeds, the performance of \NAME\ is satisfactory in controlling the three main planar degrees of freedom of the quadruped's body, namely $x$, $y$, and $\yaw$.

\subsection{Execution times test}

To verify the timing predictability of \NAME\ on an embedded system, timing experiments were conducted on a Raspberry Pi 5, including
a quad-core ARM Cortex-A76 64-bit processor, clocked at 2.4 GHz, and 16 GB of RAM, running Ubuntu 24.04 LTS.
To perform these experiments, \NAME\ was compiled into a separate C++ program, outside ROS 2.
To measure execution times as precisely as possible, the program was executed as an isolated task on a single core of the computer with maximum priority for the Linux scheduler, to prevent operating system tasks from interfering with the task executing \NAME.
To collect a significant number of measurements of the execution times for a single job, both experiments from Section~\ref{ssec:expctrl} were repeated, for a total of 23085 samples.

\begin{figure}[h]
	\includegraphics[width=\linewidth]{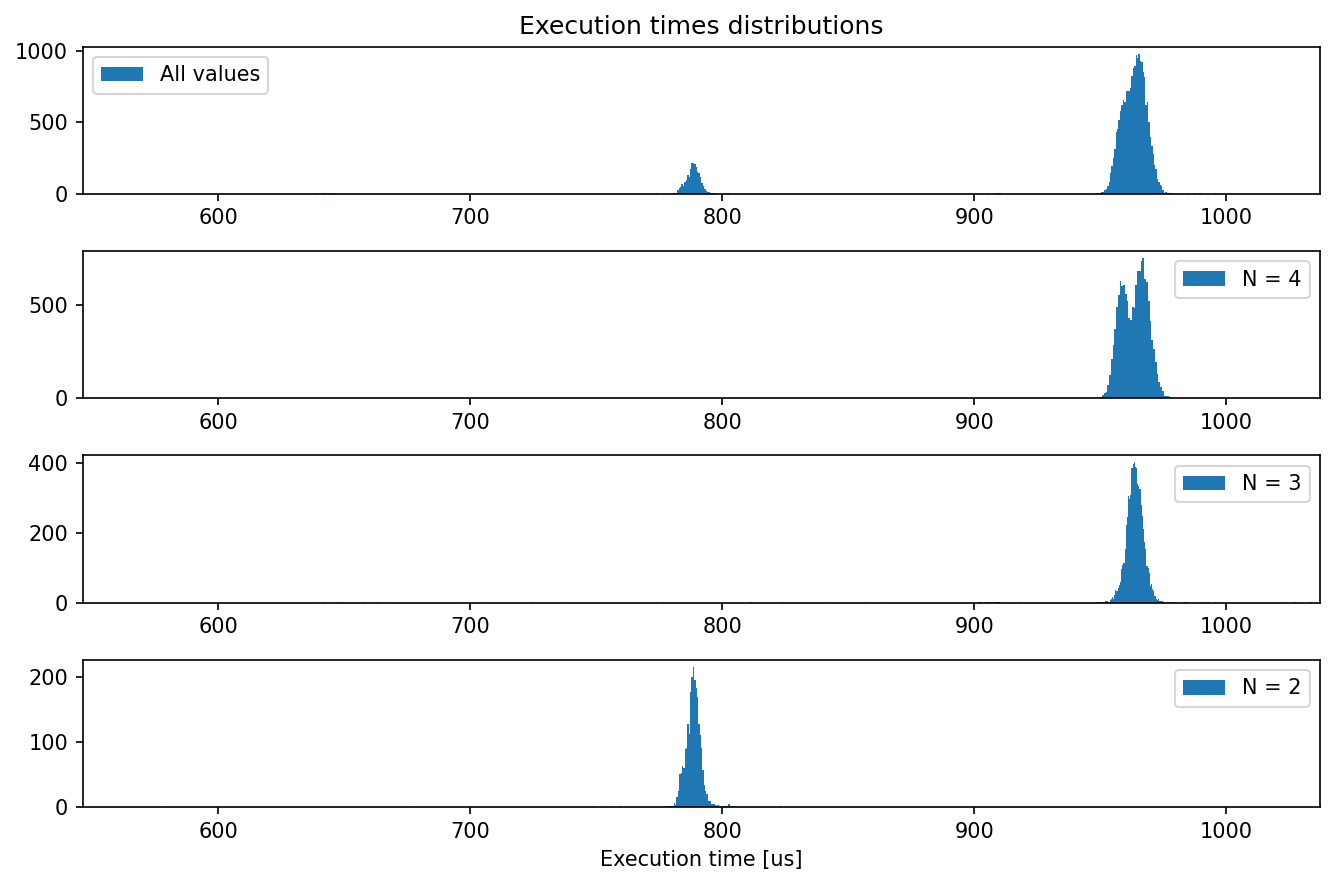}%
	\caption{Execution times distributions of the task that implements \NAME\ when a different number $N$ of feet are in contact with the ground.}%
	\label{fig:ET}
\end{figure}

Figure~\ref{fig:ET} shows the distribution of the execution times of the task that executes \NAME. In particular, the upper plot reports the overall distribution, while the other plots show the distribution when a different number $N$ of feet are in contact with the ground.

Most execution times are concentrated between 950 $\mu$s and 970 $\mu$s, when the robot has 3 or 4 grounded feet, and between 780 $\mu$s and 800 $\mu$s, when the robot has 2 grounded feet.
When the robot has 2 grounded feet, the rank $r$ of the matrix $A$ is lower compared to when the robot has 3 or 4 grounded feet, and therefore the LDQ decomposition requires fewer iterations to terminate (see Section~\ref{sssec:ldq}).
Additionally, when the robot has 2 grounded feet, the Support Polygon reduces to a segment, making it much faster to find the safe reference for the center of mass compared to when the robot has 3 or 4 grounded feet, where the Support Polygon is a triangle or a quadrilateral (see Section~\ref{ssec:bodyref}).
Moreover, the algorithm that solves Equation~\eqref{eq:OZ_safe}, in the case of 3 or 4 grounded feet, is implemented with a \textsf{for} loop containing a \textsf{break}, and this can influence the execution times, which are maximal when the desired ZMP is inside the Support Polygon.

Figure~\ref{fig:ET} shows also that the execution times for each number of legs are distributed with a seemingly unimodal probability density and a very low standard deviation.
In fact, the tails in the histograms are essentially invisible.
This makes the execution time measurements highly reliable and predictable, and therefore suitable for a real-time system.
Furthermore, the worst-case execution time (WCET) is approximately 1030 $\mu$s when there are 3 or 4 grounded feet, and approximately 820 $\mu$s when there are 2 grounded feet.
These data are crucial when scheduling tasks in a real-time system, as they allow the analysis to be performed with a low pessimism, hence, allowing to enforce timing guarantees without wasting computational resources.

	\section{Conclusions}

This paper presented \NAME, a predictable framework for quadruped locomotion designed for embedded systems with hard real-time execution constraints.

\NAME's predictability comes from the utilization of a novel closed-form control law for the body, that leverages feedback linearization, which is applied for the first time to control a quadruped robot.

Applying feedback linearization required the definition of an appropriate model for the body, enabling the system to be linearized in closed loop and inequality constraints to be transformed into references.
The model utilizes the LDQ matrix decomposition, which overcomes the under-actuation problem of quadruped robots and the loss of stability during walking, reflected in the loss of rank of the constraint matrix.
The LDQ matrix decomposition employs an algorithm that requires a fixed known number of iterations to complete, making the framework suitable for real-time execution.
Additional improvements have also been introduced in foot motion planning to make \NAME\ responsive to sudden variations in high-level input commands while maintaining accuracy and predictability.

The experiments presented in Section~\ref{sec:results} demonstrate that \NAME\ is capable of following high-level commands regardless of speed, adapting the gait to the situation. Furthermore, it can be executed on an embedded system with bounded computation times.

As a future work, we aim to extend the \textit{Feet Manager} and the \textit{Body Reference} blocks to enable the quadruped robot to handle generic uneven terrains, while always maintaining predictability.
In addition, we intend to use \NAME\ on one or more physical quadruped robots to demonstrate the validity of the framework in a real-world context.

	\bibliographystyle{ieeetr}
	\bibliography{references.bib}  
\end{document}